\documentclass{article}

\usepackage{microtype}
\usepackage{graphicx}
\usepackage{subcaption}
\usepackage{booktabs} % for professional tables

\usepackage{hyperref}

\usepackage{multirow}

\usepackage{algorithm}
\usepackage{algorithmic}

\newcommand{\ie}{\emph{i.e., }}

\newcommand{\cf}{\emph{cf. }}

\usepackage[table]{xcolor}
\definecolor{color+}{RGB}{0, 100, 0}
\definecolor{color-}{RGB}{200, 0, 0}
\definecolor{lightgray}{gray}{0.93}
\definecolor{lightred}{RGB}{255, 230, 230}
\definecolor{lightgreen}{rgb}{0.9, 1, 0.9}

\usepackage[accepted]{icml2026}

\newcommand{\slh}[1]{{\color{black}{#1}}}
\newcommand{\scs}[1]{{\color{black}{#1}}}

\newcommand{\za}[1]{{\color{black}{#1}}}
\newcommand{\concept}[1]{{\color{black}{#1}}}

\definecolor{item}{RGB}{15, 86, 157}
\definecolor{query}{RGB}{169, 16, 3}

\definecolor{-}{rgb}{0.25,0.41,0.88}
\definecolor{+}{rgb}{0.70,0.13,0.13}

\definecolor{table_color}{RGB}{239,246,251}

\usepackage[most]{tcolorbox}
\definecolor{block-gray}{gray}{0.95} % 备用灰色
\definecolor{mint-bg}{RGB}{235, 245, 240} % 仿照图片的淡青色

\newtcolorbox[auto counter]{findingbox}[1][]{
    enhanced,
    colback=mint-bg,      % 背景色
    colframe=mint-bg,     % 边框颜色（设为与背景同色即无边框感）
    boxrule=0pt,          % 边框宽度设为0
    arc=4pt,              % 圆角弧度
    left=6pt, right=6pt, top=6pt, bottom=6pt, % 内部边距
    fontupper=\bfseries Finding~\thetcbcounter:~\mdseries, % 自动加粗 "Finding X.X:"
    #1
}

\newtcolorbox{problembox}[1][]{
    enhanced,
    colback=gray!5,           % 背景色：极浅的灰色 (Gray 5%)
    colframe=gray!30,         % 边框颜色：浅灰色
    boxrule=0.5pt,            % 边框宽度：很细的线
    arc=3pt,                  % 圆角弧度
    left=6pt, right=6pt, top=6pt, bottom=6pt, % 内部边距
    fontupper=\small,         % 内容字体：稍微小一点，显得精致
    title={},                 % 无标题栏
    #1                        % 允许传入额外参数
}

\newtcolorbox{genpromptbox}[1][]{%
  colback=white,
  colframe=black,
  boxrule=1pt,
  arc=2pt,
  width=\linewidth, % 建议使用 \linewidth 适应单/双栏
  boxsep=2pt,
  left=4pt, %稍微增加一点内边距更好看
  right=4pt,
  top=4pt,
  bottom=4pt,
  title={PROMPT:}, 
  fonttitle=\large\bfseries, 
  coltitle=white, 
  #1
}

\usepackage{amsmath}
\usepackage{amssymb}
\usepackage{mathtools}
\usepackage{amsthm}
\usepackage{enumitem}

\usepackage[capitalize,noabbrev]{cleveref}

\theoremstyle{plain}

\theoremstyle{definition}

\theoremstyle{remark}

\usepackage[textsize=tiny]{todonotes}

\icmltitlerunning{Reasoning Can Be Restored by Correcting a Few Decision Tokens}

\begin{document}

\twocolumn[
  \icmltitle{Reasoning Can Be Restored by Correcting a Few Decision Tokens}
  % \icmltitle{
  % The Few that Matters: xxx
  % }



  \icmlsetsymbol{equal}{*}

  \begin{icmlauthorlist}
    \icmlauthor{Changshuo Shen}{equal,ustc}
    \icmlauthor{Leheng Sheng}{equal,nus}
    \icmlauthor{Yuxin Chen}{nus}
    \icmlauthor{An Zhang}{ustc}
    \icmlauthor{Xiang Wang}{ustc}
  \end{icmlauthorlist}

  \icmlaffiliation{ustc}{University of Science and Technology of China, Hefei, China}
  \icmlaffiliation{nus}{National University of Singapore, Singapore}

  \icmlcorrespondingauthor{An Zhang}{an\_zhang@ustc.edu.cn}

  \icmlkeywords{Large Language Models, Reasoning, Inference Intervention, Interpretability}

  \vskip 0.3in
]



\printAffiliationsAndNotice{\icmlEqualContribution}

\begin{abstract}
Large reasoning models (LRMs) substantially outperform \za{their base LLM counterparts on challenging reasoning benchmarks}, yet it remains \za{poorly understood} where base models \concept{go wrong} during token-by-token generation and how to narrow this gap efficiently.
We study the base--reasoning gap through \za{quantifying} token-level distributional disagreement between a base model and a stronger reasoning model \za{using likelihood-based divergences}.
\za{Across benchmarks,} we find that \za{\textit{\concept{the reasoning advantage is highly sparse and concentrates on a small set of early, planning-related decision tokens}}.
For instance, on Qwen3-0.6B, only $\sim$8\% of generated tokens account for the salient disagreement, and these tokens concentrate early in the response, are strongly enriched in planning-related decisions ($17\times$), and coincide with high base-model uncertainty—suggesting that base models fail mainly at early planning points that steer the subsequent reasoning trajectory.
Building on this findings, we propose disagreement-guided token intervention, a simple inference-time delegation scheme that performs a one-token takeover by the reasoning model only at high-disagreement positions and immediately switches back to the base model. 
With a small intervention budget, this sparse delegation substantially recovers and can even surpass the performance of a same-size reasoning model on challenging reasoning tasks.}
% Across multiple math-reasoning benchmarks, disagreement is highly heavy-tailed, concentrates in early-to-mid positions, aligns strongly with the base model's entropy peaks, and simple aggregates over the top-disagreement tokens predict eventual failures.
% These observations imply a sparse control view of reasoning: a few decision tokens determine whether the rollout follows a strong-model-like trajectory.
% We test this causally with an inference-time, disagreement-gated takeover that replaces the base model only at detected spikes under a token budget.
% With only a small fraction of tokens intervened, the method recovers large performance gains across reasoning benchmarks and yields trajectories that qualitatively match the stronger model.
Code is available at \url{https://github.com/AlphaLab-USTC/RRTokenIntervention}.

\end{abstract}

% Large reasoning models (LRMs) substantially outperform same-sized base models, yet it remains unclear where base models go wrong during token-by-token generation and how to narrow this gap efficiently. We study the base-reasoning gap through token-level distributional disagreement between a base model and a stronger reasoning model along the base rollout. We find that **the gap is governed by a sparse set of early, trajectory-defining disagreement spikes rather than diffuse small errors.** Across multiple math-reasoning benchmarks, disagreement is highly heavy-tailed, concentrates in early-to-mid positions, aligns strongly with the base model's entropy peaks, and simple aggregates over the top-disagreement tokens predict eventual failures. These observations imply a sparse control view of reasoning: a few decision tokens determine whether the rollout follows a strong-model-like trajectory. We test this causally with an inference-time, disagreement-gated takeover that replaces the base model only at detected spikes under a token budget. With only a small fraction of tokens intervened, the method recovers large performance gains across reasoning benchmarks and yields trajectories that qualitatively match the stronger model. Code is available at https://anonymous.4open.science/r/RRTokenIntervention-EBDD.
\section{Introduction}\label{sec:introduction}

% reasoning ability 主要体现在少量的 planning 相关的 response前期的 decision token 上。我们可以用简单的disagreement signal识别这些token， 他们和模型自身的uncertainty强相关， 并且通过在这些位置上用reasoning model来takeover 极大的恢复甚至超越其reasoning ability

\begin{figure}[t]
    \centering
    \includegraphics[width=\columnwidth]{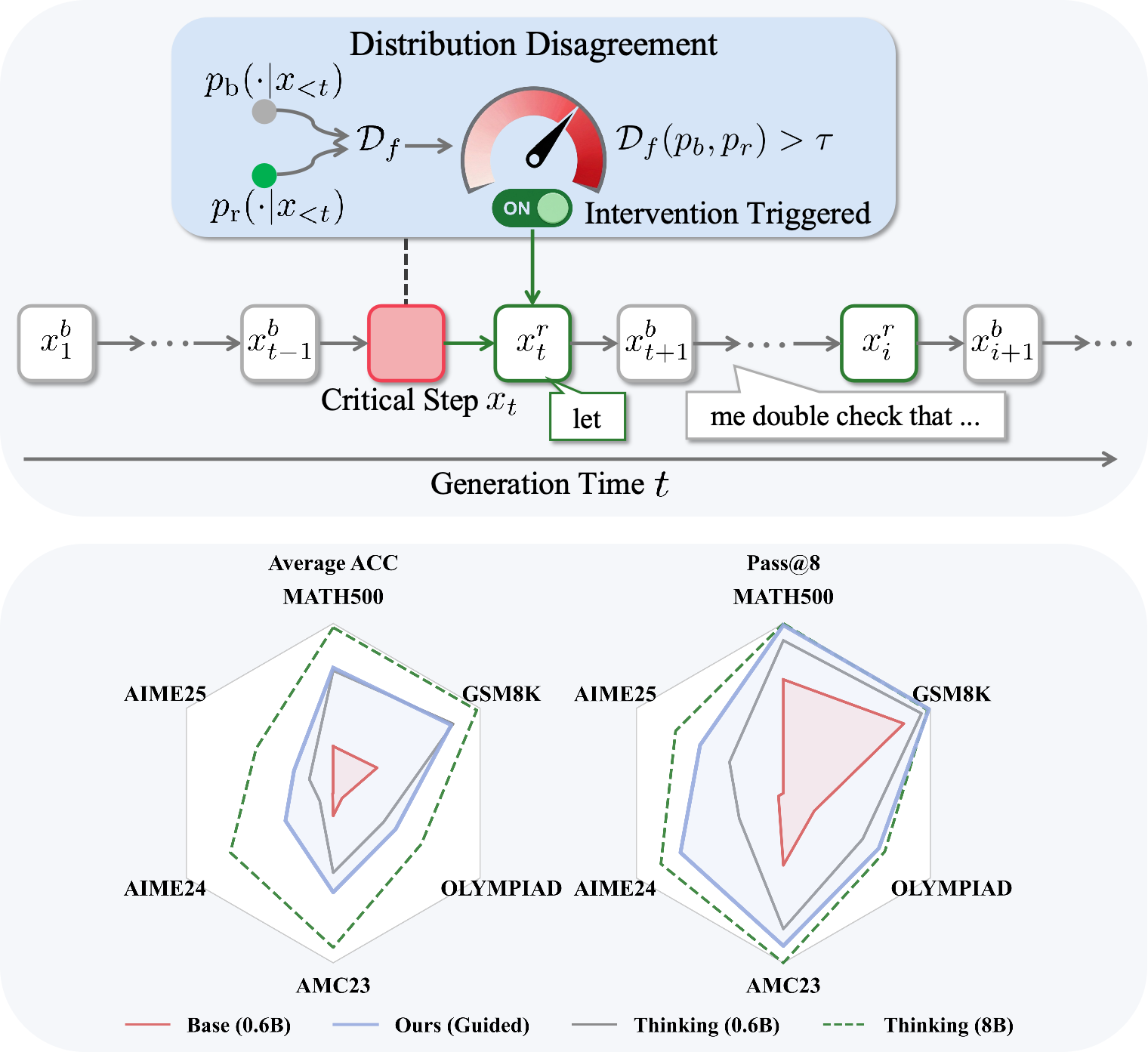}
    \vspace{-8pt}
    \caption{
    % Token-level disagreement localizes decision-critical positions and enables sparse reasoning intervention.
    % \textbf{Top:} Distributional disagreement between a base model and a reasoning model is monitored along the base rollout; large disagreement spikes trigger intervention.
    % \textbf{Bottom:} Despite intervening on a small fraction of tokens, this divergence-guided takeover recovers a substantial portion of reasoning performance across benchmarks, approaching that of full reasoning models.
    % for Qwen3-0.6B, \(\bar{\rho}\!\approx\!8\%\) already surpasses the same-size reasoning variant (\(55.4/74.6\) vs.\ \(43.4/64.1\) Avg. Acc./Pass@8).
    % Increasing the budget further bridges the gap toward the 8B reference across benchmarks. 
    \concept{Disagreement-guided token intervention} and its effectiveness. 
    % By changing few critical tokens with a strong LRM  (\ie top figure) triggered at high token-level distributional disagreement, a 0.6B base model can recover most of the reasoning capabilities of a strong 8B LRM (\ie bottom figure), and surpassing post-trained 0.6B LRM.
    \slh{
    Intervening on 8\% critical tokens, triggered by high token-level distributional disagreement through a strong LRM (\ie top figure), enables a 0.6B base model to recover most reasoning capabilities of the 8B strong LRM (\ie bottom figure) and outperform its post-trained 0.6B variant.
    }}
    \label{fig:teaser-figure}
    \vspace{-12pt}
\end{figure}

\begin{figure*}[t]
    \centering
    % ---------- Row 1: three panels ----------
    \begin{subfigure}[b]{0.33\textwidth}
        \centering
        \includegraphics[width=\textwidth]{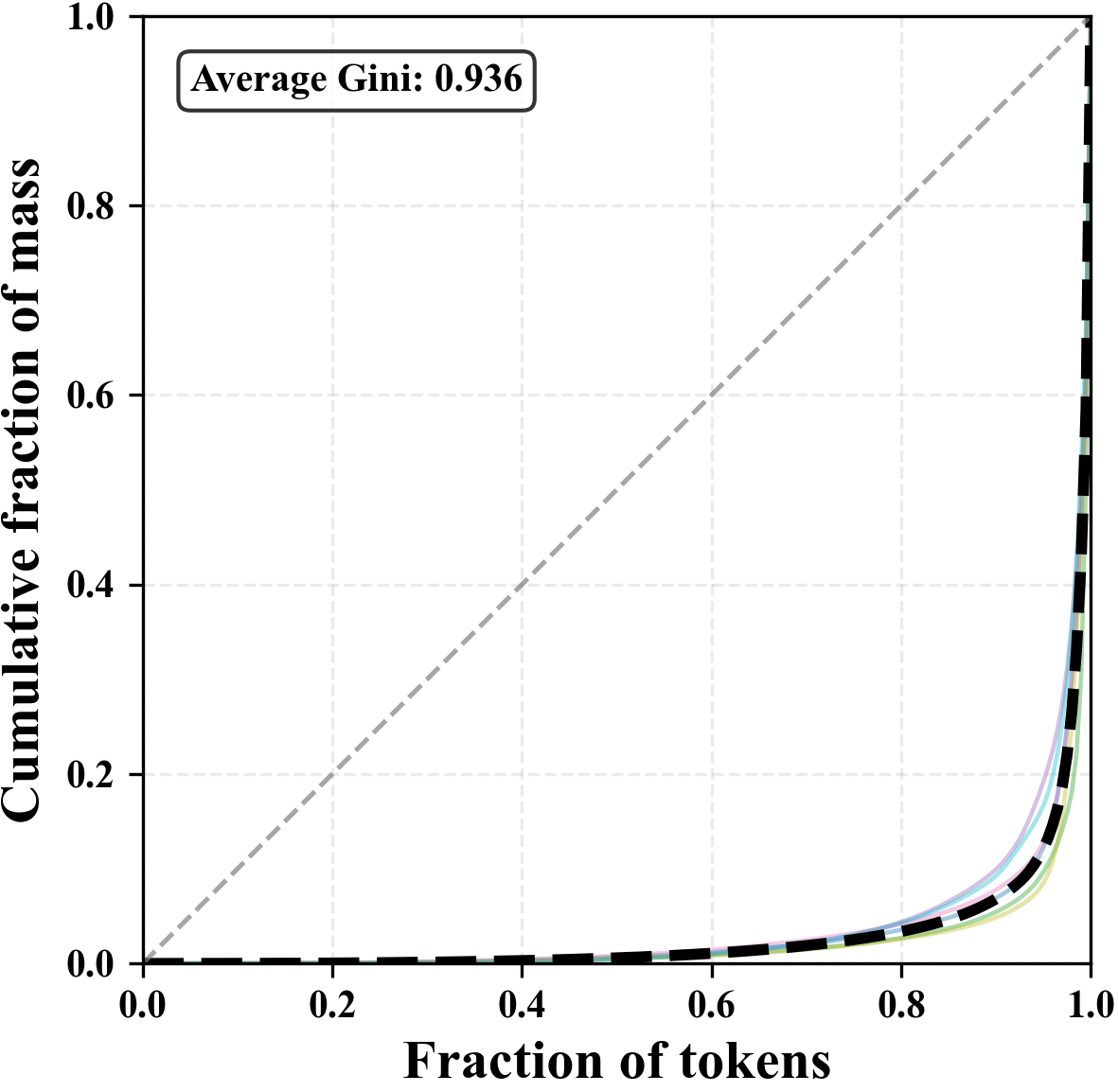}
        \caption{Disagreement is highly sparse}
        \label{fig:sparsity_lorenz}
    \end{subfigure}
    \hfill
    \begin{subfigure}[b]{0.33\textwidth}
        \centering
        \includegraphics[width=\textwidth]{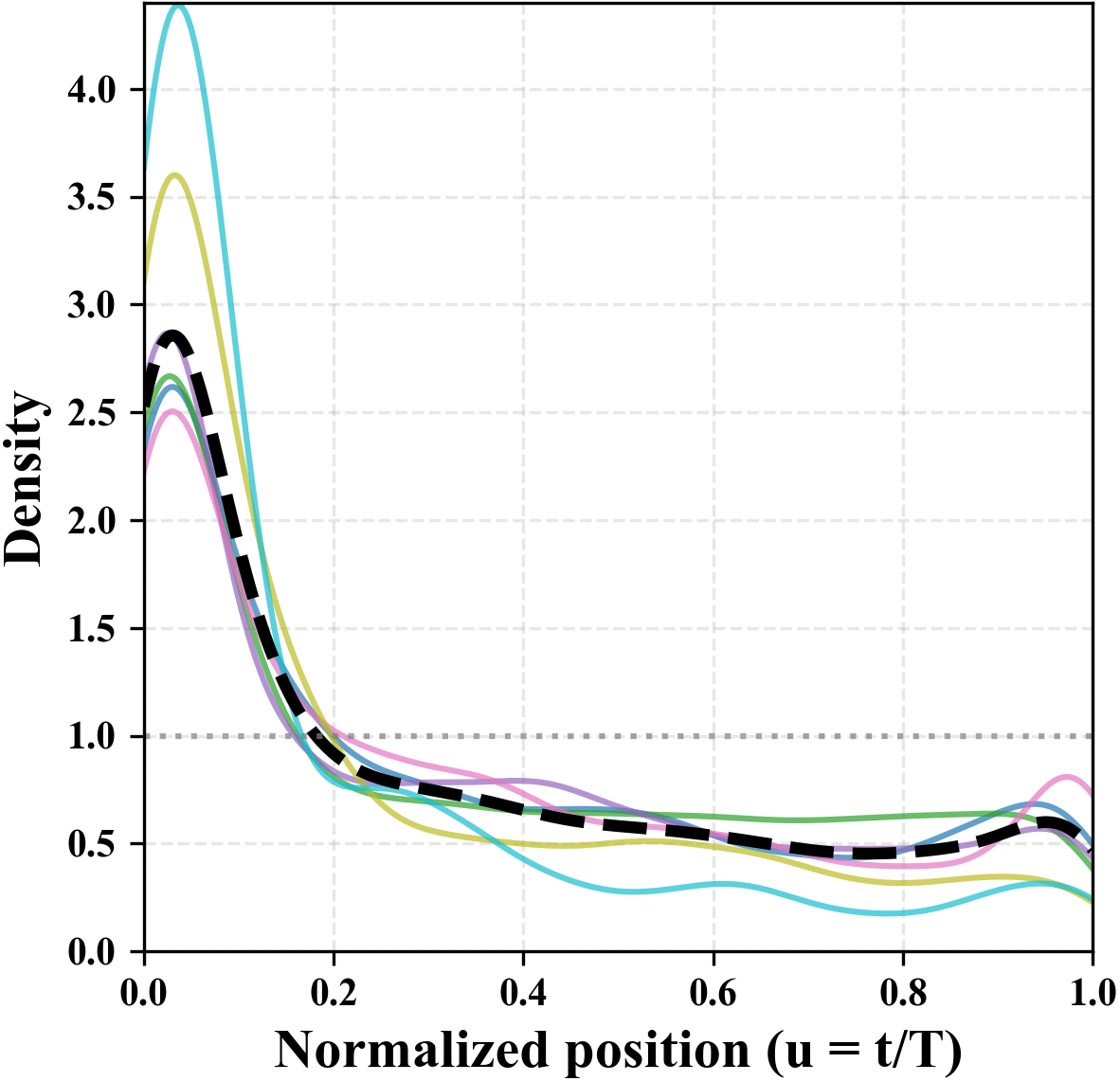} 
        % \caption{High-Disagreement Tokens Occur Early}
        \caption{High-disagreement tokens occur early}
        \label{fig:position_density}
    \end{subfigure}
    \hfill
    \begin{subfigure}[b]{0.33\textwidth}
        \centering
        \includegraphics[width=\textwidth]{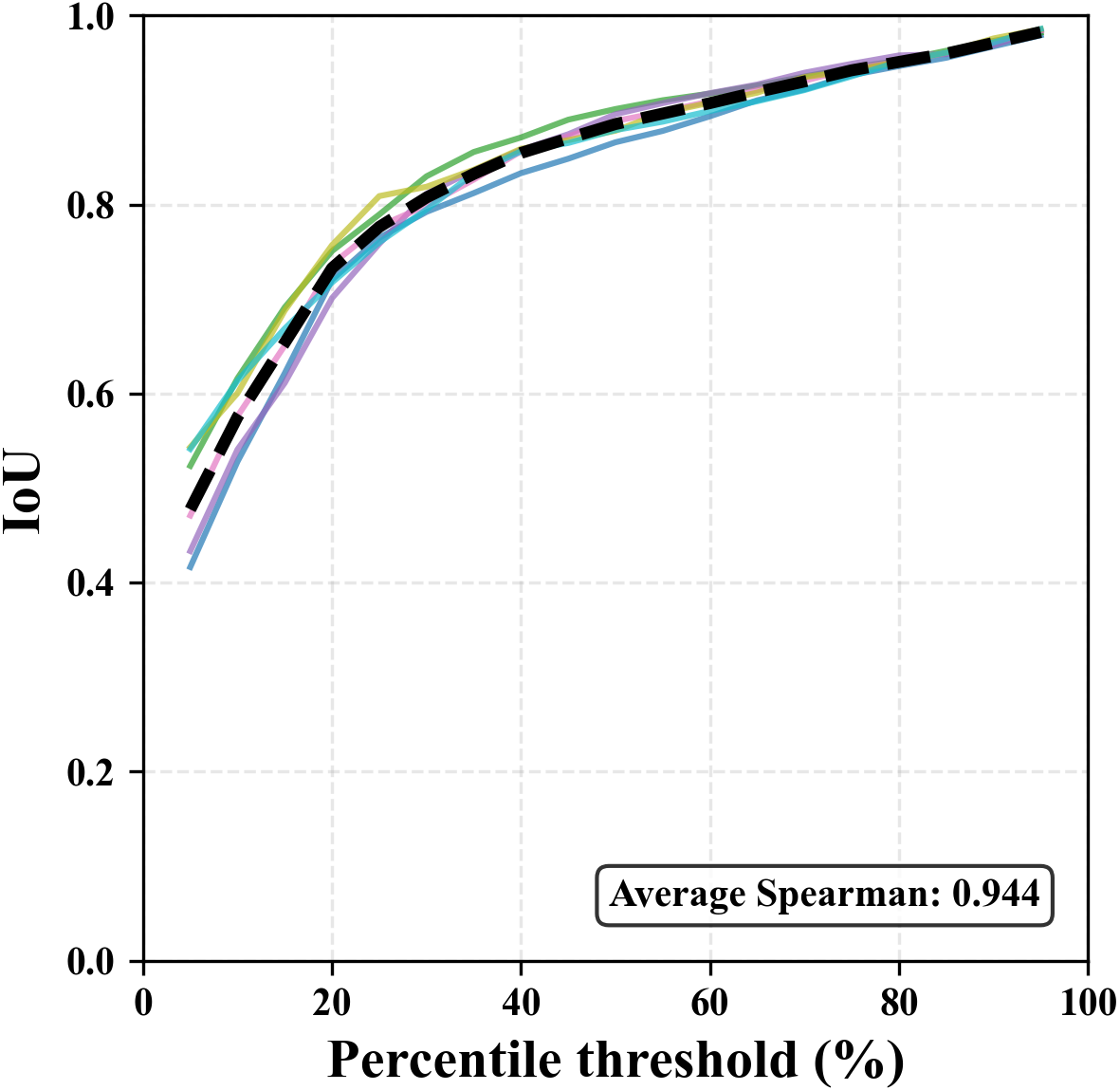} 
        % \caption{Disagreement–Uncertainty Overlap}
        \caption{Disagreement–uncertainty overlap}
        \label{fig:iou}
    \end{subfigure}
    \vspace{-8pt}
    % ---------- Row 2: shared legend ----------
    \includegraphics[width=\textwidth,trim=0 10 0 6,clip]{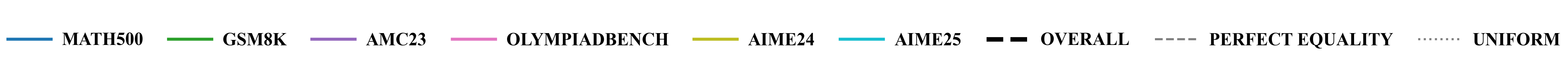}
    \vspace{-8pt}
    \caption{Token-level disagreement is sparse, early, and uncertainty-aligned across benchmarks.
    (a) Lorenz curves show that disagreement mass is highly concentrated in a small fraction of tokens.
    (b) Normalized-position density of the top-$1\%$ highest-disagreement tokens is strongly skewed toward early steps (u=t/T), indicating early-to-mid trajectory concentration relative to a uniform baseline.
    (c) IoU between top-$p\%$ disagreement tokens and top-$p\%$ entropy tokens remains high across thresholds, showing strong alignment between cross-model disagreement and base-model uncertainty.}
    \vspace{-5pt}
    \label{fig:main_diagnosis}
\end{figure*}

\begin{figure*}
    \centering
    \includegraphics[width=\linewidth]{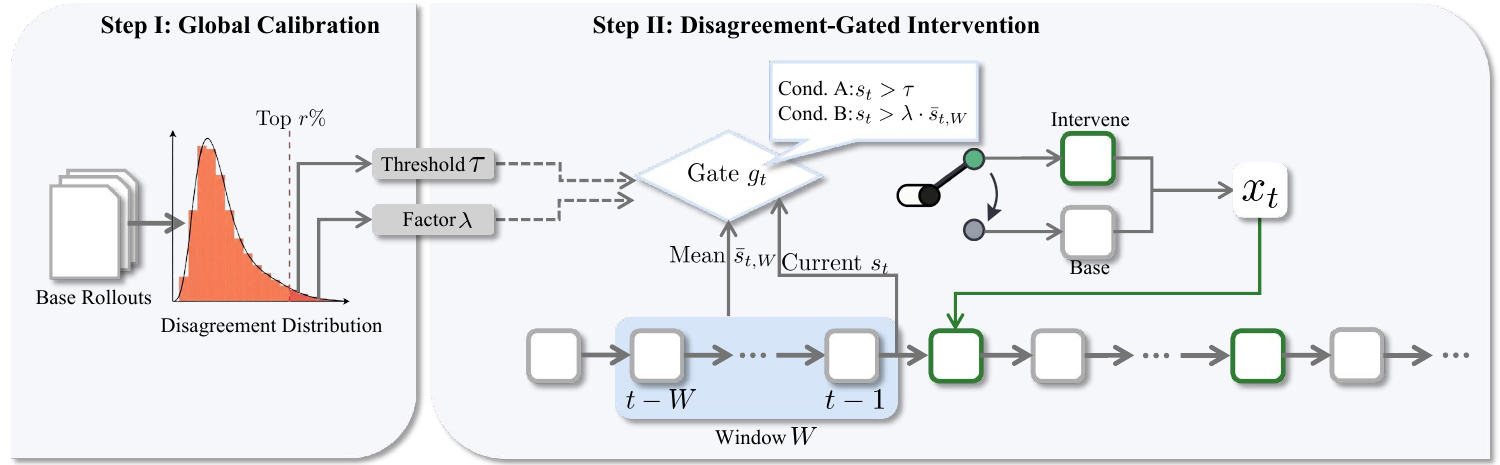}
    \caption{Illustration of Disagreement-Gated Intervention. In  {Step I}, we calibrate the disagreement threshold $\tau$ and factor $\lambda$ using the top-$r\%$ of scores from base rollouts. In  {Step II}, the gate $g_t$ dynamically switches between the Base model and the Intervene model (reasoning expert). Intervention is triggered only when the token-level disagreement $s_t$ satisfies both the global constraint ($s_t > \tau$) and the local relative spike condition ($s_t > \lambda \cdot \bar{s}_{t,W}$).}
    \vspace{-5pt}
    \label{fig:framework}
\end{figure*}

\za{
Large reasoning models (LRMs) have recently demonstrated substantial advantages over their base LLM counterparts on challenging reasoning benchmarks~\cite{survey-LLMs,openai-o1,openai-o3o4,deepseek-r1,AIME,olympaid}, with gains commonly attributed to test-time scaling~\cite{scaling-reasoning,test-time-scaling,self-consistency,ToT} and reinforcement learning-based post-training such as RLVR~\cite{deepseek-math,rlvr-incentivizing}.
Beyond further improving performance, this widening base--LRM gap has motivated growing interest in a more fundamental question: \emph{what} differs between base and reasoning models, and \emph{why} does ``reasoning mode'' work?
A growing body of evidence supports a latent-capability view in which base models may already encode considerable reasoning machinery, while post-training mainly activates, amplifies, or stabilizes it through minimal additional signals—such as activation-vector steering~\cite{actadd,reasoning-steering,representation-engineering,rl-repurposes-representation}, confidence-based self-rewarding~\cite{self-reward-lm,confidence-aware-reward-modeling}, distribution-level output amplification~\cite{echo-chamber}, or even one-example unlocks~\cite{1-shot-rl}.

To address the \emph{why} question, prior work has proposed hypotheses at multiple levels. 
Reasoning capabilities may emerge from eliciting richer intermediate trajectories, as in chain-of-thought prompting \cite{cot-prompting,least-to-most,self-consistency}; from inducing reasoning micro-behaviors such as backtracking \cite{ToT,react}, verification \cite{PRM800K}, and self-correction \cite{self-refine,survey-self-correction,deepseek-thoughtology}; from steering reasoning direction by restricting policy gradient updates to high-entropy tokens \cite{entropy-80/20}; or from updating only a small subset of parameters, \ie a top-singular reasoning subspace \cite{AlphaRL,lora}.
% Despite these insights, we still lack a simple answer for the \emph{why} question—namely, a token-level account of where reasoning ability actually manifests during generation and which decision points are truly drive the base--LRM performance gap.
Despite these insights, we still lack \concept{a simple token-level account} of \emph{why} ``reasoning mode'' works—namely, where reasoning ability actually manifests during token generation and which decision points truly drive the \concept{base--LRM} performance gap.

In this paper, we address this question by quantifying the base--LRM gap through \concept{token-level distributional disagreement} between a base model and a stronger reasoning model % its reasoning counterpart, 
using likelihood-based divergences such as cross-entropy and reverse Kullback--Leibler divergence~\cite{information-theory}.
We then characterize which tokens carry the reasoning advantage, how many such tokens exist, and what properties they exhibit. 
Across benchmarks, we find that reasoning ability is concentrated in a surprisingly small set of \concept{planning-related decision tokens} that more likely appear early in the response. 
Concretely, only a small fraction of generated tokens determines whether the model commits to a reasoning trajectory: on Qwen3-0.6B~\cite{QWEN3}, merely $\sim$8\% of output tokens account for the salient base--LRM disagreement (\ie \concept{disagreement spike}) (\cf Figure~\ref{fig:teaser-figure}), indicating that \concept{token-level disagreement} is highly sparse. 
Moreover, these disagreement spikes appear disproportionately early in generation (\cf Figure~\ref{fig:main_diagnosis}), contrary to the common intuition that reasoning gains primarily stem from long traces after the reasoning process unfolds. 
% Finally, these critical tokens are strongly associated with explicit planning: during recovery from base to reasoning rollouts, planning-related tokens become markedly more likely, increasing by $17\times$ in occurrence probability (\cf Table~\ref{tab:token_enrichment}).
% Notably, they also correlate strongly with the base model’s intrinsic uncertainty, suggesting that the most consequential decision points coincide with where the base model is least confident (\cf Table \ref{fig:iou}).
Finally, these critical tokens are tightly linked to planning under uncertainty: during recovery from base to reasoning rollouts, planning-related tokens become markedly more likely, increasing by $17\times$ in occurrence probability (\cf Table~\ref{tab:token_enrichment}), and they correlate strongly with the base model’s intrinsic uncertainty (\cf Figure~\ref{fig:iou}). 
This suggests that base models struggle on complex reasoning tasks primarily at the planning points—where they are least confident are precisely the planning decisions that steer the subsequent reasoning trajectory.

Motivated by these findings, we ask whether a base model can attain strong reasoning performance with limited assistance from a reasoning model—by delegating only a small number of critical decisions. 
We propose a simple inference-time mechanism as shown in Figure~\ref{fig:teaser-figure}: at positions where the base--LRM \concept{disagreement score} is large, we perform a one-token takeover—the reasoning model generates only the next token at that position, after which generation immediately switches back to the base model~\cite{routellm,relayllm}. 
We term this sparse delegation scheme \concept{disagreement-guided token intervention} (Figure~\ref{fig:framework}). 
Experiments show that with a small intervention budget, this approach can substantially recover—and in some settings even surpass—the performance of a same-size reasoning model on challenging reasoning tasks.
Together, these findings motivate a token-level view of reasoning: a small number of early planning commitments can steer the entire trajectory and dominate the reasoning gap.
}

\section{Related Work}\label{sec:related-work}

% \paragraph{Reasoning Capabilities in Base Models.}
% \paragraph{Reasoning capabilities in base models.}
\textbf{Reasoning capabilities in base models.}
Reasoning behaviors can emerge in small or base models when they are exposed to appropriate training signals, such as reasoning exemplars or Chain-of-Thought~\cite{cot-prompting} traces~\cite{larger-models-teach-reasoning}.
Explicit alignment, including fine-tuning on structured rationales, often suffices to elicit multi-step problem solving in models that previously answered directly~\cite{larger-models-teach-reasoning}.
Reinforcement learning has been shown to gradually restructure output patterns, increasing response length and inducing more consistent use of intermediate reasoning steps~\cite{deepseek-r1,1-shot-rl}.
Activation-space methods such as steering or targeted editing demonstrate that reasoning patterns, like backtracking, can be causally induced in base models without weight updates~\cite{reasoning-steering,rl-repurposes-representation}.
Inference-time approaches, such as CoT prompting or self-verification feedback, can also trigger reasoning behaviors that do not naturally surface in zero-shot settings~\cite{cot-prompting,congnitive-bahavior4reasoning}.
Collectively, these findings suggest that base models possess a wide range of reasoning capabilities that can be activated through post-training mechanisms, guidance signals, or structural interventions.

% Many recent works reveal that base models know how to reason, post training helps them how to correctly deploy these reasoning capabilities\cite{base-models-reason, congnitive-bahavior4reasoning, deepseek-thoughtology, echo-chamber, rl-repurposes-representation, 1-shot-rl}. % 这种句子都要在结尾句号前面cite

% 再想一下，*信息量*
% \paragraph{Token-Level Distributional Disagreement.} 
% \paragraph{Token-level distributional disagreement.} 
\textbf{Token-level distributional disagreement.}
Cross-entropy, entropy, and divergence offer effective signals for identifying decision-critical positions where models diverge or hesitate during generation.
% Token-level metrics that quantify distributional divergence or uncertainty provide useful signals for identifying decision-critical positions in generation. 
These signals can help locate forking tokens, error-prone steps, or places where weak models diverge sharply from stronger references.
\textit{KL divergence}, in both forward and reverse form,has been used to compare different models' behavior. 
\textit{Forward KL} penalizes the student for deviating from the teacher’s distribution, while \textit{Reverse KL} emphasizes tokens where the student is overconfident in outputs disfavored by the teacher~\cite{distillation,distill-minillm,distill-opd1}.
% highlights tokens where the student model assigns high confidence to outputs that the teacher model deems unlikely, which has been used for distillation~\cite{distill-minillm} especially on policy distillation~\cite{distill-opd1,distill-opd2}.
\textit{Entropy} measures local uncertainty in the model's predictive distribution. 
Entropy spikes often correspond to reasoning forks or hesitation. 
Focusing updates on high-entropy tokens improves RLVR efficiency~\cite{entropy-80/20}, while entropy-guided rollout branching improves exploration~\cite{fr3e}.
\textit{Cross-entropy} measures the magnitude of divergence between a model’s output and a target distribution, and is widely used as a training loss.
In contrast, \textit{$\Delta \log p$}, the difference between log probabilities in model before and after post-training,  captures the directionality of model updates.
Such metrics support the identification of high-impact tokens where reasoning behavior changes or supervision is most needed.

% \paragraph{Selective Strong Model Interventions in Inference} 
% \paragraph{Selective intervention at inference time.}
\textbf{Selective intervention at inference time.}
% aim to enhance weak-model generation by inserting strong-model control without replacing the entire rollout.
The reasoning capabilities of weak models can be enhanced by inserting strong-model control without replacing the entire rollout.
These approaches typically fall into two paradigms: collaborative inference for efficiency and guided generation for distillation stability.
In collaborative inference, strong models serve as on-demand verifiers to balance performance and compute.
At the request level, expensive resources can be reserved for hard queries by dispatching inputs based on predicted capability gaps~\cite{routellm}.
For finer-grained control, verification mechanisms allow strong models to screen and take over from lightweight drafters upon detecting quality degradation~\cite{cascade-sd}.
Furthermore, difficult reasoning steps can be explicitly ``relayed'' to a strong model~\cite{relayllm} or escalated to the cloud~\cite{adaswitch-cloud-local} specifically when small models exceed internal risk thresholds.
Beyond inference acceleration, selective interventions are crucial for stabilizing on-policy distillation by preventing student models from learning from erroneous trajectories~\cite{distillation,distill-minillm,distill-opd1}.
To ensure data quality, teacher distributions can serve as filters to reject unlikely tokens~\cite{sdk-distillation}, or dynamically take over generation at divergence peaks to correct reasoning paths~\cite{adaswitch-distillation}.
Together, these strategies demonstrate that concentrating teacher assistance on sparse, high-impact positions effectively bridges the gap between weak and strong models in both inference and training stages.

\section{Preliminary}\label{sec:preliminary}

% task scope
% inputs/outputs
% structural property
% objective
% constrains

% \slh{In this paper, we mainly study the behavioral differences from the token-level generation perspective between the base model and a stronger reasoning model, where two models share the same vocabulary. Here we briefly introduce the model generation process and how to measure token-level disagreement during the generation process.}

In this paper, we focus on the token-level behavioral differences between a standard base model and a stronger reasoning model sharing the same vocabulary.
In this section, we establish the framework for our analysis.
We first introduce the comparative generation setup and the metrics for quantifying distributional disagreement in \cref{sec:setup}.
Then, we formally formulate the problem of sparse reasoning takeover in \cref{sec:problem_formulation}.

\subsection{Setup and Notations}\label{sec:setup}

% 增加Motivation，为什么要设定这两个模型
% \textbf{Autoregressive Generation Process.}
\textbf{Autoregressive generation process.}
To characterize the reasoning gap, we consider a standard autoregressive generation setting involving two models sharing a vocabulary $\mathcal{V}$: a base model $\mathcal{M}_b$ and a stronger reasoning model $\mathcal{M}_r$~\cite{attention,gpt3,distillation,speculative-decoding}.
Given a prompt $x_0$, the model generates a sequence $y = (y_1, \dots, y_T)$.
At each step $t$, let $x_t = (x_0, y_{<t})$ denote the context.
We denote the next-token probability distributions as $p_b(\cdot \mid x_t)$ and $p_r(\cdot \mid x_t)$ respectively.
Typically, $\mathcal{M}_r$ exhibits superior reasoning capabilities (e.g., via larger scale or specialized post-training) but incurs higher computational costs, whereas $\mathcal{M}_b$ is efficient but performance-limited~\cite{scaling-law,rlhf,cot-prompting}.
Our goal is to understand how $\mathcal{M}_b$ deviates from the superior trajectory of $\mathcal{M}_r$.

% \textbf{Metrics for disagreement.}
\slh{
\textbf{Quantified disagreement metrics.}}
To quantify the behavioral disagreement between the two models at a granular level, we define a generic \concept{token-level disagreement score} $s_t$ along the trajectory generated by the base model.
Specifically, at step $t$ with context $x_t$, the disagreement is measured by a disagreement function $\mathcal{D}_f$ between the base model's distribution $p_b(\cdot \mid x_t)$ and the reasoning model's distribution $p_r(\cdot \mid x_t)$:
\begin{equation}
s_t = \mathcal{D}_f \big(p_b(\cdot \mid x_t),\, p_r(\cdot \mid x_t)\big)
% \mathcal{D}(t) \;=\; f\big(p_b(\cdot \mid x_t),\, p_r(\cdot \mid x_t)\big).
\end{equation}
In this work, we primarily utilize the Cross Entropy ($\mathrm{CE}$) as the disagreement metric~\cite{information-theory}.
It measures the expected surprisal of the reasoning model with respect to the base model's choices:
\begin{equation}
\begin{aligned}
\mathcal{D}_f = \mathcal{D}_{\mathrm{CE}}(t)
&= H(p_b(\cdot \mid x_t), p_r(\cdot \mid x_t))\\
&= -\sum_{y\in\mathcal{V}} p_b(y\mid x_t) \log p_r(y\mid x_t).
\end{aligned}
\end{equation}

We note that other metrics, such as the Reverse KL divergence, are also applicable and closely related (i.e., $D_{\mathrm{rKL}} = \mathcal{D}_{\mathrm{CE}} - H_b$).
These metrics serve as the primary probing signals to localize where the base model deviates from the reasoning model's decision path.
In parallel, we monitor the base model’s intrinsic uncertainty using the Shannon entropy $H_b(t) = H(p_b(\cdot \mid x_t))$.

\subsection{Problem Formulation}\label{sec:problem_formulation}

While the reasoning model $p_r$ outperforms the base model $p_b$, deploying it for every token generation is computationally prohibitive.
Our objective is to bridge this performance gap by intervening only on a sparse subset of tokens.
Formally, we aim to identify a set of \textit{decision-critical positions} $\mathcal{S} \subset \{1, \dots, T\}$, such that $|\mathcal{S}| \ll T$.

We define a \concept{disagreement-guided token intervention} mechanism where the generation follows the base model $p_b$ by default, but switches to the reasoning model $p_r$ at indices $t \in \mathcal{S}$.
This can be viewed as a dynamic interplay where the base model handles execution tokens, and the reasoning model intervenes solely at decision tokens.

Our core hypothesis is that the performance gap is not uniformly distributed across all tokens but is concentrated at positions of high distributional disagreement.
Therefore, the identification of $\mathcal{S}$ can be formulated as a detection problem using the disagreement metrics defined in Section~\ref{sec:setup}.
If $D_{\mathrm{CE}}(t)$ effectively captures these critical branching points, we can recover the reasoning capabilities of $\mathcal{M}_r$ with a minimal intervention budget.
In the following sections, we will first analyze the distribution of $D_f$ to validate this sparsity hypothesis (Section~\ref{sec:diagnosis}), and then propose a concrete algorithm for budgeted intervention (Section~\ref{sec:intervention}).
% 这两个标题需要换成观点性的句子
% \section{Base--Reasoning Differences Concentrate at Sparse, Trajectory-Defining Steps}
% \section{Analysis of Token-Level Base--LRM Gap}
\slh{
\section{Token-Level Base--LRM Disagreement}
% \section{A Few Early Disagreement Spikes Define the Reasoning Trajectory}
\label{sec:diagnosis}

% 这里主要是做一些统计分析
    % 稀疏性：base-reasoning的差异不是diffuse、而是集中在少数position
    % 结构性：这些position具有"decision tokens"的结构特征，比如早中期、分叉、规划语意等
    % 预测性：对最终失败具有预测能力

% In this section, we provide a statistical diagnosis of the token-level generation differences between base and reasoning models.
% Our goal is to operationalize the reasoning gap by identifying where exactly the divergence occurs.
% We validate three core hypotheses:
% (i) cross-model disagreement is {sparse} and heavy-tailed;
% (ii) high-disagreement positions exhibit {structure} (e.g., earlier planning stages) and align with the base model's {uncertainty};
% and (iii) aggregated disagreement is  {predictive} of eventual reasoning failures.
In this section, we provide a comprehensive analysis of the token-level distributional disagreement between base models and LRMs. 
Specifically, we have the following core findings:
% (i) cross-model disagreement is {sparse} and heavy-tailed;
(i) token-level distributional disagreement happens at few token positions, which is {sparse} and heavy-tailed (\cf Section \ref{subsec:sparsity});
% (ii) high-disagreement positions exhibit {structure} (e.g., earlier planning stages) and align with the base model's {uncertainty};
(ii) token-level distributional disagreement tends to happen at early positions (\cf Section \ref{subsec:structure});
(iii) token-level distributional disagreement tends to align with the uncertainty of the base model, concentrating on planning-related tokens (\cf Section \ref{subsec:alignment});
% and  exhibit {structure} (e.g., earlier planning stages) and align with the base model's {uncertainty};
and (iv) token-level distributional disagreement degree can be utilized for predicting the final reasoning failures (\cf Section \ref{subsec:predictive}). }
% aggregated disagreement is  {predictive} of eventual reasoning failures.

% Our analysis primarily utilizes the
We conduct analysis mainly on the {Qwen3} family~\cite{QWEN3}, employing the  {Qwen3-0.6B-Base} model for rollout generation ($p_b$) and the  {Qwen3-8B} (Thinking) for reasoning guidance ($p_r$).
Evaluations are conducted across six mathematical benchmarks ranging from standard tasks ( {GSM8K}~\cite{GSM8K},  {MATH500}~\cite{MATH}) to competition-level challenges ( {AIME}~\cite{AIME},  {OlympiadBench}~\cite{olympaid},  {AMC23}~\cite{AMC23}).
Detailed configurations for models, datasets, and baselines are provided in Appendix~\ref{apdx:experimental-setup}.
We follow the setup in \cref{sec:setup} and use token-level disagreement $s_t$ (Cross Entropy by default) computed along the base rollout.

% \begin{figure}[t]
%     \centering
%     \includegraphics[width=\columnwidth]{figures/part1/gini.png}
%     \caption{ {Compact sparsity summary.} 
%     (Top) Gini coefficients of $s_t$ across datasets. 
%     (Bottom) Top-5\% share of disagreement mass.}
%     \label{fig:gini_topk_bar}
% \end{figure}

% \subsection{Disagreement is Sparse and Heavy-Tailed}
% \subsection{Disagreement Concentrates in Rare Spikes}
\slh{
\subsection{Disagreement Concentrates in Few Token Positions}
\label{subsec:sparsity}
}

\begin{findingbox}
% The vast majority of disagreement mass is concentrated in a tiny fraction of  {high-disagreement tokens}.
Disagreement mass is extremely concentrated: a tiny fraction of tokens accounts for most cross-model disagreement.
\end{findingbox}

We study the per-token disagreement scores $\{s_t\}$ along base-model rollouts and ask whether the gap reflects diffuse, incremental drift or a small number of spikes.
\slh{
Figure~\ref{fig:sparsity_lorenz} shows that token-level distributional disagreement is {sparse} and heavy-tailed, since the Lorenz curve lies far below the equality line (\ie most tokens accumulate little mass of the disagreement while a small subset dominates). }
% indicating that most tokens contribute little while a small subset dominates~\cite{Lorenz}. }

% Figure~\ref{fig:sparsity_lorenz} shows that disagreement mass is highly uneven: 
% the Lorenz curve bends sharply below the equality line, indicating that most tokens contribute negligible mass while a small subset dominates~\cite{Lorenz}.

We summarize this concentration with the Gini coefficient, which quantifies how unevenly the cumulative disagreement mass is distributed across token positions  (0 = uniform, 1 = maximally concentrated, \cf Appendix~\ref{apdx:lorenz_gini})~\cite{gini,Lorenz}. 
Across six math reasoning benchmarks, we obtain an average $G=0.936$ (details in Appendix~\ref{apdx:lorenz_gini}).
This extreme inequality supports a sparse control view: the base--reasoning gap is driven primarily by rare, high-magnitude deviations at a small set of trajectory-sensitive positions, rather than by many small errors accumulating over time. 
% Here we refer the as 

% \subsection{High-Disagreement Tokens Occur Early}
% \subsection{Disagreement Spikes Cluster Early}
\slh{
\subsection{Disagreement Spikes Happen at Early Positions}\label{subsec:structure}
}

\begin{findingbox}
High-disagreement spikes are front-loaded: most occur in the earliest portion of generation rather than being spread uniformly across the rollout.
% Disagreement peaks are not uniformly distributed but are structurally skewed toward the early stages of generation.
\end{findingbox}

We examine the positional structure of disagreement by normalizing each token index as $u=t/T\in[0,1]$ and selecting, for each sample, the top-1\% tokens by disagreement score $s_t$. 
We then aggregate these selected positions across samples and estimate their density.
As shown in Figure~\ref{fig:position_density}, the resulting distribution is strongly left-skewed: spikes occur very early (mode at $u\!\approx\!0.05$) and the density rapidly decays, becoming close to baseline by $u\!\approx\!0.2$.

This pattern is inconsistent with a view where reasoning gains primarily come from long traces after the reasoning process unfolds.
Instead, because early Chain-of-Thought tokens often encode decomposition and strategy selection~\cite{cot-prompting}, the positional skew supports a planning gap interpretation: the models diverge most at early trajectory-setting commitments, not during later execution.

% \subsection{Disagreement Spikes Track Uncertain Planning Forks}
\slh{
\subsection{Disagreement Spikes Relate to Model Uncertainty}\label{subsec:alignment}
}

\begin{table}[t]
    \centering
    \caption{\textbf{Planning enrichment at uncertainty/disagreement spikes.}
    We label tokens as \textit{planning} vs.\ \textit{execution} and report the share of planning tokens in the global token stream versus among top-$1\%$ tokens ranked by either base entropy $H_b(t)$ or cross-model divergence $s_t$.
    \textit{Enrich.} is the ratio of the two shares.}
    \label{tab:base_planning_enrichment}
    \setlength{\tabcolsep}{3pt}
    \begin{tabular}{lcccc}
        \toprule
        Metric & Global (\%) & Top-$p\%$ (\%) & Enrich. \\
        \midrule
        Entropy $H_b(t)$ & 1.89 & 15.75 & 8.31$\times$ \\
        Disagreement $s_t$ & 1.89 & 14.13 & 7.46$\times$ \\
        \bottomrule
    \end{tabular}
\end{table}

\begin{findingbox}
Disagreement spikes are not arbitrary: they coincide with base-model uncertainty spikes and are strongly enriched for planning-related tokens.
% Moments of high cross-model disagreement strongly coincide with the base model's uncertainty peaks, which are enriched for planning-related tokens.
\end{findingbox}

\textbf{Disagreement spikes coincide with intrinsic uncertainty.}
We ask whether cross-model disagreement $s_t$ highlights structurally meaningful positions or merely reflects superficial stylistic differences.
Figure~\ref{fig:iou} reports the Intersection-over-Union (IoU) between tokens ranked by disagreement score $s_t$ (CE-based) and those ranked by the base model entropy $H_b(t)$, computed per sample over the top-$p\%$ sets and averaged.
The strong overlap shows that disagreement spikes concentrate at steps where the base model is intrinsically uncertain, providing an internal validation that these spikes correspond to difficult, trajectory-sensitive points rather than random mismatch.

% 是不是可以直接dsiagreement spikes are planning rich
\textbf{These uncertainty-linked disagreement spikes are planning-rich.}
We further characterize the content of these positions.
Using a lightweight heuristic classifier for \textit{planning} vs.\ \textit{execution} tokens (Appendix~\ref{apdx:enrichment_details}), we compute planning enrichment among the top-1\% tokens ranked by $s_t$ and $H_b(t)$ respectively within base-model rollouts~\cite{enrichment-analysis}.
As shown in Table~\ref{tab:base_planning_enrichment}, planning markers are strongly over-represented: across all datasets, the planning share rises from 1.89\% globally to 15.75\% at entropy spikes (8.31$\times$) and to 14.13\% at disagreement spikes (7.46$\times$).
Appendix~\ref{apdx:base_planning_enrichment_per_dataset} provides per-dataset results, confirming the trend is consistent across benchmarks.

Taken together, disagreement spikes systematically track uncertain planning commitments---positions where a small change in planning-related tokens can steer the subsequent trajectory.

\slh{
\subsection{Disagreement Spikes Predict Reasoning Failures}\label{subsec:predictive}
}

\begin{figure}[t]
    \centering
    \includegraphics[width=\columnwidth]{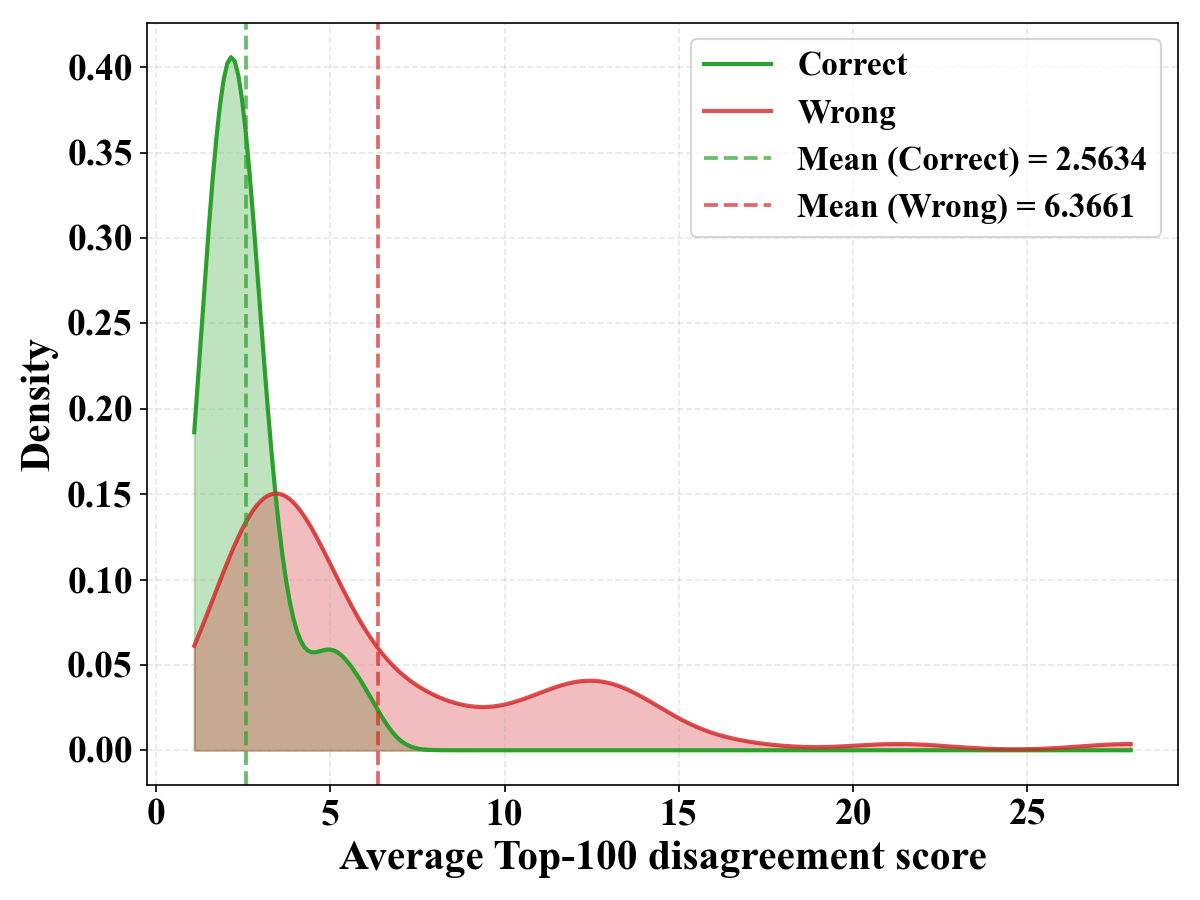}
    \caption{The distribution of top-100 mean disagreement scores shows a distinct separation, where incorrect samples (red) exhibit significantly higher values than correct ones (green).}
    \label{fig:predict-failure}
    \vspace{-10pt}
\end{figure}

\begin{findingbox}
Aggregating disagreement over a small set of high-disagreement tokens predicts reasoning failures. 
\end{findingbox}

The previous analyses suggest that disagreement is (i) extremely concentrated in rare spikes, (ii) front-loaded early in the rollout, and (iii) aligned with uncertain, planning-heavy steps.
If these spikes indeed reflect trajectory-setting mistakes, then their magnitude within a sample should be predictive of whether the rollout ultimately succeeds.

% Finally, we test the predictive power of these sparse signals.
% For each sample, we compute a scalar failure score as the mean of the top-100 disagreement values $s_t$ (ranked within the sample) and evaluate AUROC for predicting incorrect answers; we use the same aggregation for entropy.
% As shown in \cref{tab:auroc}, on GSM8K, Disagreement achieves an AUROC of  {0.851} compared to Entropy's 0.817.
% These results indicate that sparse disagreement signals are  {predictive} of eventual failures, motivating a causal test via targeted intervention in the next section.

\begin{table}[t]
    \centering
    \caption{ {Failure prediction performance.} We compare failure prediction using base model uncertainty (Entropy) versus our cross-model disagreement (Cross Entropy), aggregated as the  {mean of top-100 tokens per sample}.}
    \label{tab:auroc}
    \resizebox{\columnwidth}{!}{% 如果表格稍微有点宽，可以用这个缩小，或者去掉这一层直接用 tabular
    \begin{tabular}{@{}llcc@{}}
        \toprule
        Dataset & Metric (Top-100 Mean) & AUROC $\uparrow$ & Acc $\uparrow$ \\
        \midrule
        \multirow{2}{*}{ {MATH500}} 
          & Entropy (Baseline) & 0.718 & 0.650 \\
          &  {Disagreement (Ours)} &  {0.773} &  {0.710} \\
        \midrule
        \multirow{2}{*}{ {GSM8K}}   
          & Entropy (Baseline) & 0.817 & 0.760 \\
          &  {Disagreement (Ours)} &  {0.851} &  {0.840} \\
        \bottomrule
    \end{tabular}
    }
\end{table}

\textbf{Failure score from disagreement spikes.}
For each sample, we form a scalar score by averaging the top-$K$ ($K=100$) per-token disagreement values $s_t$ within that sample.
We evaluate how well this score predicts incorrect final answers using AUROC and Accuracy; we compute an analogous score using base entropy $H_b(t)$ for comparison.

As shown in Table~\ref{tab:auroc} and Figure~\ref{fig:predict-failure}, the disagreement-based score provides a clear separation between correct and incorrect samples and consistently outperforms entropy (e.g., GSM8K: 0.851 vs.\ 0.817 AUROC).
This result closes the loop with the preceding structure findings: the same sparse, early, planning-linked spikes that dominate disagreement mass also carry actionable, sample-level predictive signal.

These observations motivate a question: if failure is driven by a small set of high-disagreement decisions, can selectively correcting only those positions recover strong-model reasoning under a small token budget?
In the next section, we answer this question by introducing a disagreement-guided, token-level intervention at inference time.

\begin{table*}[t]
    \centering
    \small
    \setlength{\tabcolsep}{3pt}
    \caption{Experimental results on mathematical reasoning benchmarks (Accuracy / Pass@8). The Recovery column quantifies how much the intervention recovers reasoning capabilities, calculated as $(P_{Guided} - P_{Base}) / (P_{Thinking} - P_{Base})$, where $P$ denotes Pass@8.}
    \label{tab:main_performance}
    \begin{tabular}{l ccccccc c}
        \toprule
        Model & Math500 & GSM8K & AMC23 & Olympiad & AIME24 & AIME25 & Average & Recovery\\
        \midrule
        Qwen3-0.6B-Base        & 27.6 / 67.0 & 29.9 / 82.0 & 13.4 / 42.5 & 6.8 / 21.0 & 0.4 / 3.3   & 0.0 / 0.0   & 13.0 / 36.0 & - \\
        \rowcolor{lightred}
        ~ + Guided ($\bar{\rho} \approx 0.04$) & 44.6 / 89.0 & 56.5 / 96.0 & 35.9 / 77.5 & 20.2 / 49.0 & 10.0 / 36.7 & 7.5 / 20.0 & 29.1 / 61.4 & $91\%$ \\
        \rowcolor{lightred}
        ~ + Guided ($\bar{\rho} \approx 0.13$) & 73.9 / 99.0 & 80.4 / 99.0 & 58.4 / 90.0 & 42.5 / 65.0 & 32.5 / 70.0 & 26.7 / 56.7 & 52.4 / 80.0 & $157\%$ \\
        \rowcolor{lightgray}
        Qwen3-0.6B (Thinking)  & 72.1 / 90.0 & 81.6 / 94.0 & 46.9 / 80.0 & 34.3 / 54.0 & 9.2 / 30.0  & 16.1 / 36.7 & 43.4 / 64.1 & - \\
        \midrule
        Qwen3-1.7B-Base        & 54.2 / 82.0 & 75.4 / 93.0 & 28.1 / 65.0 & 19.2 / 41.0 & 5.0 / 20.0   & 2.5 / 10.0   & 30.8 / 51.8 & - \\
        \rowcolor{lightred}
        ~ + Guided ($\bar{\rho} \approx 0.06$) & 66.8 / 95.0 & 72.0 / 97.0 & 50.6 / 95.0 & 33.2 / 64.0 & 20.4 / 46.7 & 18.8 / 43.3 & 43.6 / 73.5 & $76\%$ \\
        \rowcolor{lightred}
        ~ + Guided ($\bar{\rho} \approx 0.16$) & 81.5 / 100.0 & 88.9 / 98.0 & 71.9 / 97.5 & 48.9 / 67.0 & 45.0 / 73.3 & 36.2 / 66.7 & 62.1 / 83.8 & $112\%$ \\
        \rowcolor{lightgray}
        Qwen3-1.7B (Thinking)  & 88.9 / 100.0 & 90.9 / 97.0 & 80.0 / 95.0 & 52.5 / 63.0 & 39.2 / 73.3  & 32.8 / 53.3 & 64.1 / 80.3 & - \\
        \midrule
        Qwen3-8B (Thinking)   & 97.4 / 100.0 & 97.8 / 98.0 & 90.9 / 100.0 & 60.0 / 69.0 & 70.0 / 83.3 & 52.5 / 73.3 & 78.1 / 87.3 & - \\
        \bottomrule
    \end{tabular}
\end{table*}
% 8%和13.3%

% \section{Correcting Sparse Disagreement Spikes Recovers Strong-Model Reasoning}\label{sec:intervention}

\slh{
\section{Reasoning Can Be Restored by Correcting a Few Decision Tokens}\label{sec:intervention}
}

% \section{Restoring Reasoning by Fixing Decision Tokens}\label{sec:intervention}

\cref{sec:diagnosis} shows that base--reasoning disagreement is 
(i) extremely sparse and mostly happens at early positions of the trajectory,
(ii) aligned with the base model's intrinsic uncertainty and enriched for planning markers,
and (iii) predictive of failure.
Motivated by this sparse-control picture, we conduct an intervention test: 
can selectively correcting only a small set of disagreement spikes yield large performance recovery under a small replacement budget?

\begin{findingbox}
A disagreement-gated, token-level replacement at a small set of spikes is sufficient to redirect rollouts and recover a disproportionate fraction of strong-model reasoning performance under small budgets.
\end{findingbox}

% \subsection{Disagreement-Gated Token Intervention}\label{subsec:takeover}

\subsection{Calibrated Disagreement Spikes for Gated Intervention}
\label{subsec:takeover}

We design a minimal inference-time intervention that alters only a small subset of emitted tokens, leaving model parameters and hidden states unchanged.
At each decoding step $t$, we compute a \concept{token-level disagreement score}
$
s_t = \mathcal{D}_f\!\left(p_b(\cdot\!\mid\!x_t),\,p_r(\cdot\!\mid\!x_t)\right),
$
where $\mathcal{D}$ is a distributional disagreement between next-token distributions of the base and reasoning models.
% Unless otherwise stated, we instantiate $\mathcal{D}$ with cross entropy (CE), consistent with \cref{sec:diagnosis}.
Unless otherwise stated, we instantiate $\mathcal{D}$ with cross entropy (CE), consistent with \cref{sec:diagnosis}; we report results with reverse-KL as an alternative disagreement metric in Appendix~\ref{apdx:rkl_results}.

\textbf{Global calibration via a spike ratio $r$ (budget control).}
To avoid per-task tuning, we calibrate a global spike criterion on a held-out calibration set by defining disagreement spikes as the top-$r$ tail of disagreement scores.
Concretely, let $\mathcal{S}=\{s_t\}$ denote all disagreement scores collected along base rollouts on the calibration prompts.
We set the global threshold $\tau$ as the $(1-r)$-quantile of $\mathcal{S}$:
\begin{equation}
\tau = Q_{1-r}(\mathcal{S}),
\end{equation}
so that $s_t>\tau$ corresponds to a top-$r$ disagreement event under the calibration distribution, and the calibrated thresholds transfer robustly across benchmarks (Appendix~\ref{apdx:calibration_sensitivity}).
We additionally compute a tail-to-mean scale factor
\begin{equation}
\lambda = \frac{\mathbb{E}[s \mid s>\tau]}{\mathbb{E}[s]},
\label{eq:lambda_def}
\end{equation}
which summarizes how much larger disagreement scores are within the top-$r$ tail compared to the global average.

\textbf{Runtime gating.}
During decoding, we trigger takeover only when $s_t$ forms a robust spike under a two-part gate:
\begin{equation}
g_t = \mathbb{I}\Big[
s_t > \tau
\ \wedge\
s_t > \lambda \cdot \overline{s}_{t,W}
\Big],~
\overline{s}_{t,W}=\frac{1}{W}\sum_{i=1}^{W} s_{t-i},
\label{eq:gate_rule}
\end{equation}
where the global threshold $s_t>\tau$ enforces a calibrated tail event, and the sliding-window ratio test suppresses noisy toggling by requiring the current score to exceed the recent local baseline by a factor $\lambda$.
Token generation then follows
\begin{equation}
y_t \sim
\begin{cases}
p_r(\cdot\mid x_t), & g_t=1,\\
p_b(\cdot\mid x_t), & g_t=0.
\end{cases}
\label{eq:takeover_rule}
\end{equation}
We also report the realized intervention rate $\rho$, the fraction of tokens generated by the reasoning model:
$\rho = \frac{1}{T}\sum_{t=1}^T g_t$.
The per-problem realized rate $\rho_i$ is concentrated with small variance across problems (Appendix~\ref{apdx:budget_stability}).

\textbf{Remark on practical deployment.}
The disagreement gate queries both $p_b$ and $p_r$ at every step and is best viewed as a diagnostic form of sparse control rather than an optimized inference system.
A cheaper alternative gates on the base entropy $H_b(t)$ alone, which already recovers most of the strong-model Pass@8 at small budgets (Appendix~\ref{apdx:entropy_variant}).

\begin{figure}[t]
    \centering
    \includegraphics[width=0.48\textwidth]{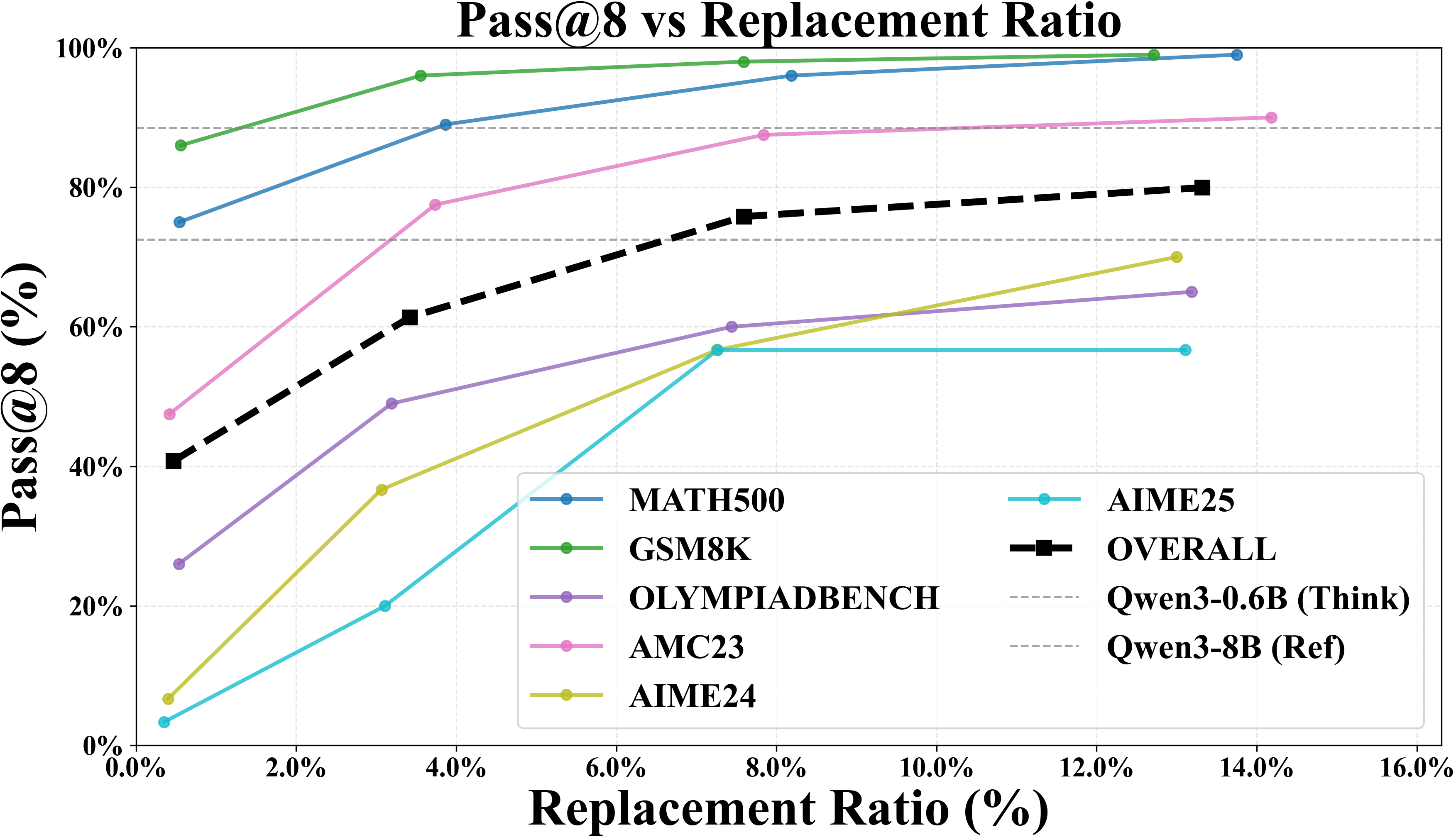}
    \vspace{-5pt}
    \caption{Performance recovers sharply with few replacement.}
    \label{fig:performance_vs_replace_ratio}
    \vspace{-8pt}
\end{figure}

\begin{figure*}
    \centering
    \begin{subfigure}[t]{0.46\textwidth} 
        \centering
        \vspace{0pt}
        \includegraphics[width=\textwidth]{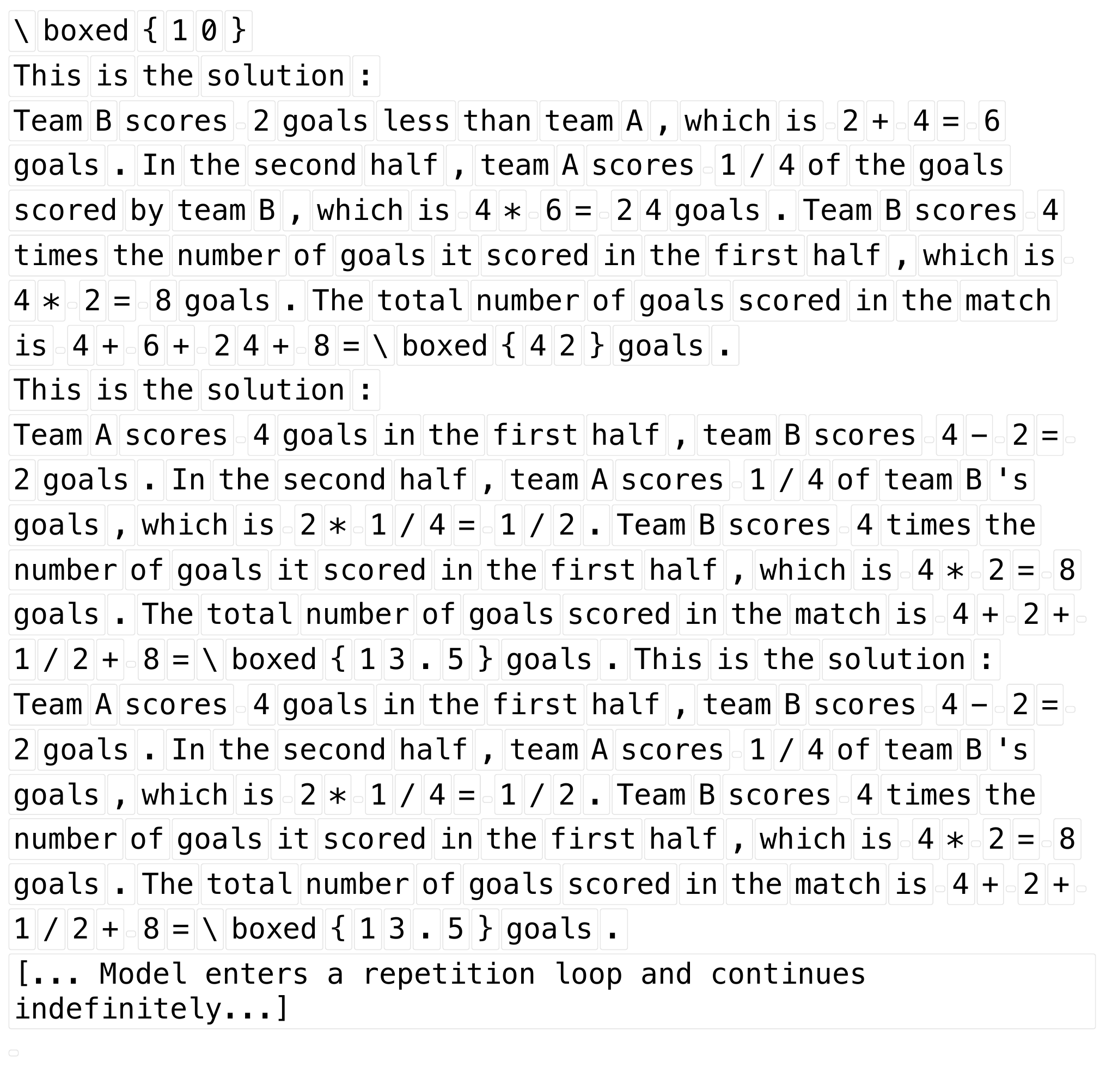}
        \caption{Base: brittle parsing and degeneration}
        \label{fig:base-response}
    \end{subfigure}
    \hfill % 撑开间距
    \begin{subfigure}[t]{0.48\textwidth}
        \centering
        \vspace{0pt}
        \includegraphics[width=\textwidth]{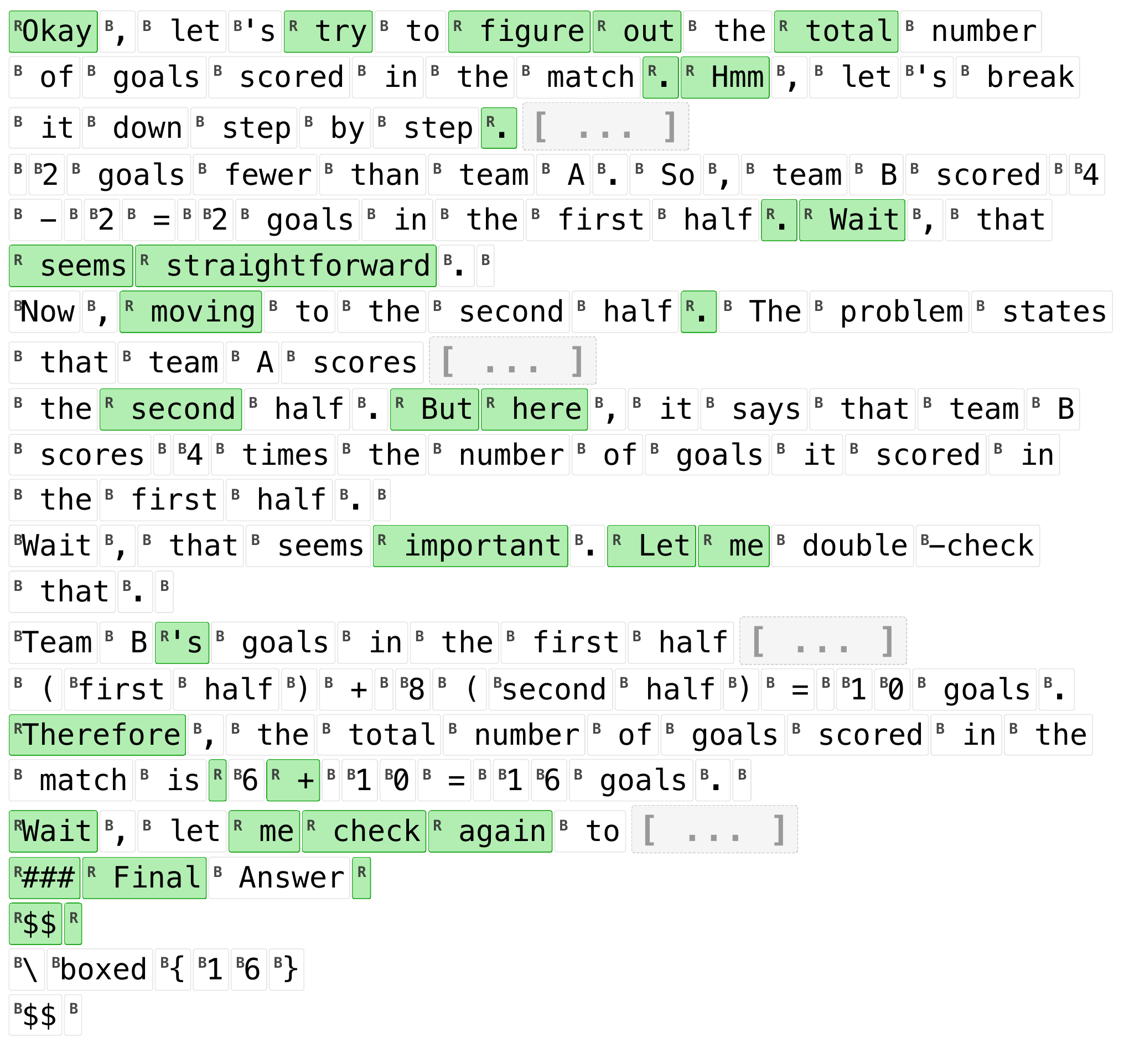}
        \caption{Guided: brief stop-and-check resolves the ambiguity}
        \label{fig:guided-response}
    \end{subfigure}
    % \vspace{-20pt}
    \caption{\textbf{Qualitative pattern: takeover induces a ``stop-and-check'' planning step.}
    Green highlights tokens generated under takeover (reasoning model); white tokens are generated by the base model.
    In this example, a short self-check triggered at an ambiguous clause fixes the variable binding, after which control returns to the base model for routine execution.}
    \label{fig:case_study}
    \vspace{-8pt}
\end{figure*}

\subsection{Sparse Interventions Yield Disproportionate Performance Recovery}\label{subsec:recovery}

We evaluate how much reasoning performance can be restored by intervening only at disagreement spikes.
Because a fixed spike ratio $r$ induces different takeover frequencies across prompts, we report results against the  {realized} intervention rate
$\rho=\frac{1}{T}\sum_{t=1}^T g_t$.

\textbf{A small budget can outperform a same-size reasoning model.}
Table~\ref{tab:main_performance} shows that sparse takeover is highly effective.
For Qwen3-0.6B, guided decoding at $\bar{\rho}\!\approx\!0.13$ improves the average from $13.0/36.0$ to $52.4/80.0$ (Acc./Pass@8), surpassing the same-size thinking variant ($43.4/64.1$) and recovering $157\%$ of its Pass@8 gap.
Qwen3-1.7B exhibits the same behavior: at $\bar{\rho}\!\approx\!0.16$, guided decoding reaches $62.1/83.8$, essentially matching its thinking counterpart ($64.1/80.3$; $112\%$ recovery).
A sample-level analysis further confirms that the intervention is rarely harmful: across $400$ problems, it flips $152$ examples from incorrect to correct and only $3$ in the reverse direction (Appendix~\ref{apdx:harm_analysis}).

\textbf{Recovery is sharply nonlinear in replacement ratio.}
Figure~\ref{fig:performance_vs_replace_ratio} reveals a pronounced knee: most of the recovery arrives within the first few percent replacement, followed by clear diminishing returns.
% \textbf{Recovery is sharply nonlinear in budget.}
% Figure~\ref{fig:performance_vs_replace_ratio} shows a steep early gain: replacing a small fraction of tokens yields most of the improvement, with diminishing returns as $\rho$ increases.
Concretely, the overall Pass@8 jumps by roughly \(\sim\!20\) points within the first few percent replacement (from \(\approx 41\%\) at \(\rho\!\approx\!0.5\%\) to \(\approx 61\%\) at \(\rho\!\approx\!3\%\)), while later increases in \(\rho\) deliver progressively smaller gains (saturating near \(\approx 80\%\) by \(\rho\!\approx\!13\%\)).
Notably, around \(\rho\!\approx\!7\%\!-\!8\%\) the overall curve already reaches the Qwen3-0.6B (Thinking) reference line, and approaching the 8B reference with larger budgets.

\textbf{Position selection, not strong-model injection, drives the gain.}
A natural concern is whether the recovery merely reflects that any injection of strong-model tokens helps.
We rule this out with two matched-budget baselines: a random baseline that uniformly samples replacement positions, and an early-only baseline that replaces the first $\rho\!\cdot\!L_r$ tokens (motivated by Finding~2's early skew).
As shown in Table~\ref{tab:baselines_main}, both baselines at an aggressive $\rho=0.25$ are dominated by guided intervention at only $\bar{\rho}\!\approx\!0.04$---a $\sim$$6\times$ smaller budget.
This shows that the gain stems from which tokens are corrected rather than from invoking the strong model alone, and that position alone is not sufficient: within the early region, only the disagreement spikes are truly trajectory-setting.
Full per-budget curves are provided in Appendix~\ref{apdx:baselines}.

\begin{table}[t]
    \centering
    \footnotesize
    \setlength{\tabcolsep}{5pt}
    \caption{\textbf{Matched-budget baselines (Qwen3-0.6B).}
    Average Acc / Pass@8 over six math benchmarks. Guided intervention at $\bar{\rho}\!\approx\!0.04$ exceeds both baselines at $\rho=0.25$.}
    \label{tab:baselines_main}
    \begin{tabular}{lccc}
        \toprule
        Method & Budget $\rho$ & Avg Acc & Avg Pass@8 \\
        \midrule
        Base                & 0.00                       & 13.0 & 36.0 \\
        ~ + Random          & 0.25                       & 26.4 & 55.2 \\
        ~ + Early-only      & 0.25                       & 25.7 & 58.4 \\
        \rowcolor{lightred}
        ~ + Guided (Ours)   & $\bar{\rho}\!\approx\!0.04$ & \textbf{29.1} & \textbf{61.4} \\
        \bottomrule
    \end{tabular}
\end{table}

\textbf{Generalization beyond Qwen3.}
The same sparse-control behavior transfers to a LLaMA pair (LLaMA-3.1-8B as base, DeepSeek-R1-Distill-Llama-8B as guide), where intervening on $\sim$20\% of tokens recovers $\sim$91\% of the Pass@8 gap (Appendix~\ref{apdx:llama_results}).
The calibration is also robust: thresholds re-fitted on each benchmark stay within the same order of magnitude and preserve the recovery curve (Appendix~\ref{apdx:calibration_sensitivity}).

Taken together, these results show that the base--reasoning gap can be closed largely by correcting a sparse set of high-disagreement decisions: a few-percent replacement already matches or exceeds the size-matched thinking model.

\subsection{Intervention Induces Reasoning Patterns}\label{subsec:patterns}

\begin{table}[t]
    \centering
    % \small
    \caption{\textbf{Enrichment Analysis.} We compare the token distribution in the global generation vs. our intervention set. Planning tokens are effectively targeted.}
    \label{tab:token_enrichment}
    \setlength{\tabcolsep}{3.5pt}
    \begin{tabular}{lccc}
        \toprule
        Category & Global (\%) & Interv. (\%) & Enrich. \\
        \midrule
        \textbf{Planning} & 1.9 & 33.3 & \textbf{17.6$\times$} \\
        Execution & 98.1 & 66.7 & 0.7$\times$ \\
        \bottomrule
    \end{tabular}
    \vspace{-15pt}
\end{table}

Beyond final accuracy, we ask  {what} the gate intervenes on and  {how} those interventions change the trajectory.
We find a consistent functional pattern: takeover is concentrated on planning markers and often manifests as a brief stop-and-check step that resolves an upstream ambiguity before handing control back to the base model.

\textbf{Takeover targets planning rather than execution.}
Table~\ref{tab:token_enrichment} quantifies the token types in our takeover set.
Although planning tokens constitute only 1.9\% of classifiable tokens in base rollouts, they account for 33.3\% of takeover tokens (17.6$\times$ enrichment).
Conversely, execution tokens are under-represented (0.7$\times$), suggesting that the gate largely trusts the base model for routine arithmetic once the high-level structure is fixed.

\textbf{A representative example: induced stop-and-check planning.}
Figure~\ref{fig:case_study} illustrates how this specialization appears in generation.
The base model fails to stably bind variables under a syntactically ambiguous clause and degenerates.
Under takeover, the reasoning model injects a short self-check that disambiguates the dependency, after which decoding returns to the base model for straightforward computation.
We provide the full prompt and additional qualitative examples in Appendix~\ref{apdx:case_study}.
This planning-specialized takeover provides a mechanistic account for the sharp budget--performance curve in Figure~\ref{fig:performance_vs_replace_ratio}: a small number of corrected planning commitments can steer long downstream execution.
\section{Limitations}\label{sec:limitations}

% 1. 我们的intervention是从findings角度走的，没有经过优化所依efficiency不高
% 2. 由于计算量开销，我们希望在更大size的model和更多reasoning benchmark上进行验证
This work has several limitations. 
% First, due to compute constraints, our study mainly uses the Qwen family and focuses on math-centric reasoning benchmarks.
% We aim to extend the token-level disagreement diagnosis and spike-guided takeover to additional model families, broader model sizes and other reasoning domains.
First, due to compute constraints, our study primarily focuses on the Qwen3 family and math-centric reasoning benchmarks; we provide initial evidence that the same sparse-control phenomenon transfers to a LLaMA-based pair (Appendix~\ref{apdx:llama_results}) and to non-math scientific QA on GPQA-Diamond (Appendix~\ref{apdx:gpqa_results}), but broader coverage of model families, model sizes, and reasoning domains (e.g., code, multi-hop QA, agentic reasoning) remains future work.
Second, as a diagnostic-first findings paper, we adopt a simple, transparent intervention rule and do not aim to exhaustively optimize system efficiency or compete with heavily engineered baselines.

% This work has several limitations. 
% First, our experiments primarily rely on the Qwen model family. While effective, it remains to be verified whether our findings on sparse disagreement generalize to other reasoning-dense architectures, such as Llama. 
% Second, our evaluation focuses mainly on mathematical reasoning benchmarks like GSM8K and MATH. 
% We have not yet explored whether the proposed intervention method is equally effective in other reasoning-intensive domains, such as code generation.

% \section{Conclusion}\label{sec:conclusion}

% This paper investigated the nature of the performance gap between base and reasoning models through the lens of token-level distributional disagreement. 
% We showed that this gap is governed by a sparse set of early, trajectory-defining spikes rather than diffuse errors, strongly aligning with the base model's intrinsic uncertainty. 
% We introduced a disagreement-gated takeover mechanism, demonstrating that intervening on a minimal budget of tokens causally recovers a disproportionate amount of reasoning performance. 
% Qualitatively, we revealed that these sparse interventions effectively decouple high-level planning from low-level execution, inducing sophisticated strategies like ambiguity resolution. 
% Our work supports a sparse control view of reasoning, offering a pathway for efficient inference where models selectively upgrade decision quality only at pivotal branching points.

\section{Conclusion}\label{sec:conclusion}

We investigated the base--reasoning gap through \concept{token-level distributional disagreement}.
Across reasoning benchmarks, disagreement highly concentrates in rare spikes that cluster early, align with base-model uncertainty, and are enriched for planning-related decision tokens; aggregating these spikes further predicts eventual failures.
Motivated by these findings, we introduced a globally calibrated, disagreement-gated token intervention that edits only a small subset of tokens.
Under modest realized takeover budgets, this intervention recovers a disproportionate fraction of strong-model reasoning performance and can outperform a same-size reasoning variant.
Qualitatively, intervention primarily targets planning rather than execution.
Mechanistically, because generation is autoregressive, correcting a few early planning-related tokens redirects the subsequent trajectory while leaving routine execution to the base model.
Overall, our results support a sparse control view of reasoning: a few early planning commitments can steer long downstream trajectories, enabling selective upgrades only at branching points.

% \clearpage
\section*{Impact Statement}

\slh{
This paper studies the token-level distributional disagreement between base models and large reasoning models (LRMs), and shows that token-level interventions can help the base model largely recover the LRM’s reasoning capability. 
Generally, this paper may yield a positive societal impact by helping us understand the underlying mechanism of LLM reasoning.
}

% \section*{Acknowledgements}
% We thank the anonymous reviewers and the area chair for their constructive feedback that helped improve this paper.

\bibliography{icml2026}
\bibliographystyle{icml2026}

% \newpage
\appendix
\onecolumn

\section{Experimental Setup} \label{apdx:experimental-setup}
\subsection{Implementation Details} \label{apdx:implementation-details}

\paragraph{Compute Environment and Models.}
We implement all experiments in PyTorch\footnote{\url{https://pytorch.org/}} using the Hugging Face Transformers library, and rely on vLLM~\citep{vllm} for efficient, high-throughput inference.
All evaluations are conducted on a single compute node equipped with an NVIDIA A100 GPU.
To enable a controlled comparison between base generation and reasoning guidance (with a shared tokenizer and vocabulary), we use the Qwen3 model family~\cite{QWEN3}.
Specifically, we use \texttt{Qwen3-0.6B-Base} as the primary rollout generator ($\mathcal{M}_b$) for its efficiency, and \texttt{Qwen3-8B} as the stronger reference model ($\mathcal{M}_r$), which provides disagreement scores and takeover tokens when interventions are triggered.

\paragraph{Prompting and Inference Configuration.}
We wrap each problem using the following template and employ sampling-based decoding with temperature $0.6$ and top-$p$ $0.95$.
Unless otherwise specified, we generate $K=8$ samples per problem in parallel.
To accommodate long chain-of-thought traces in competition-level mathematics, we set the maximum generation length to 16{,}384 tokens.

\begin{center}
\begin{genpromptbox}
Please reason step by step, and put your final answer within \textbackslash boxed\{\}.
\\ This is the problem:
\\ \{\{prompt\}\}
\end{genpromptbox}
\end{center}

\paragraph{Sampling Protocol and Metrics.}
For each problem, we draw $K=8$ independent samples under the same decoding configuration.
We report \textit{Pass@8} as whether  {any} of the 8 samples is verified as correct.
For \textit{Accuracy}, we report the  {average} correctness across the same 8 samples (i.e., the mean of 8 Bernoulli outcomes), rather than greedy decoding.

\paragraph{Answer Verification.}
We verify predicted answers using the \texttt{math\_verify} library.
Concretely, we parse both the gold answer and the model response using a joint extraction configuration (\texttt{ExprExtractionConfig} and \texttt{LatexExtractionConfig}), and then apply \texttt{verify(gold, pred)} to test mathematical equivalence.
If parsing fails or the verifier raises an exception (e.g., due to unsupported SymPy objects), we conservatively mark the prediction as incorrect.

\paragraph{Intervention Mechanism and Calibration.}
Our disagreement-gated intervention is grounded in Cross Entropy (CE) between the next-token distributions of the base and reference models.
Intervention thresholds are calibrated on a held-out set of 500 prompts randomly sampled from the GSM8K training split~\cite{GSM8K}, ensuring no overlap with any test set.
As reported in our results, we evaluate two primary sparsity settings, $r=0.03$ and $r=0.05$, targeting the top 3\% and 5\% of disagreement spikes, respectively.
To improve stability and reduce spurious triggers, we apply sliding-window smoothing (see \cref{eq:gate_rule}) with window size $W=64$, intervening only when the current disagreement significantly exceeds the local moving average.

\paragraph{Pseudocode.}
\cref{alg:disagreement_gate} provides a self-contained pseudocode for the disagreement-gated intervention.
The calibration step (lines 1--4) is run once on a held-out set; the runtime step (lines 5--14) is invoked online during decoding.
The procedure does not modify any model parameters or hidden states.

\begin{algorithm}[h]
\caption{Disagreement-Guided Token Intervention}
\label{alg:disagreement_gate}
\textbf{Input:} base model $p_b$, reasoning model $p_r$, calibration prompts $\mathcal{C}$, target spike ratio $r$, sliding-window size $W$, max length $T_{\max}$, prompt $x_0$. \\
\textbf{Output:} response $y_{1:t}$.
\begin{algorithmic}[1]
\STATE \textit{// --- Phase I: Offline calibration (run once) ---}
\STATE Run base rollouts on $\mathcal{C}$ and collect $\mathcal{S}=\{s_t\}$ with $s_t=\mathrm{CE}(p_b(\cdot\!\mid\!x_t),\,p_r(\cdot\!\mid\!x_t))$
\STATE $\tau \leftarrow Q_{1-r}(\mathcal{S})$ \hfill \textit{// $(1-r)$-quantile threshold}
\STATE $\lambda \leftarrow \mathbb{E}[s\mid s>\tau] / \mathbb{E}[s]$ \hfill \textit{// tail-to-mean scale}
\STATE \textit{// --- Phase II: Online decoding for prompt $x_0$ ---}
\STATE Initialize $y_{1:0}\leftarrow\emptyset$
\FOR{$t = 1, \dots, T_{\max}$}
    \STATE $x_t \leftarrow (x_0, y_{1:t-1})$
    \STATE Compute $s_t = \mathrm{CE}(p_b(\cdot\!\mid\!x_t),\,p_r(\cdot\!\mid\!x_t))$
    \STATE $\overline{s}_{t,W} \leftarrow \frac{1}{\min(t-1,\,W)}\sum_{i=1}^{\min(t-1,\,W)} s_{t-i}$ \hfill \textit{// skip gate if $t=1$}
    \STATE $g_t \leftarrow \mathbb{I}[s_t > \tau \,\wedge\, s_t > \lambda\cdot\overline{s}_{t,W}]$
    \IF{$g_t = 1$}
        \STATE $y_t \sim p_r(\cdot\!\mid\!x_t)$ \hfill \textit{// one-token takeover}
    \ELSE
        \STATE $y_t \sim p_b(\cdot\!\mid\!x_t)$
    \ENDIF
    \IF{$y_t = \texttt{<EOS>}$}
        \STATE \textbf{break}
    \ENDIF
\ENDFOR
\STATE \textbf{return} $y_{1:t}$
\end{algorithmic}
\end{algorithm}

\clearpage
\subsection{Evaluation Benchmarks} \label{apdx:benchmarks}

% We evaluate the proposed method and baselines on a comprehensive suite of mathematical reasoning benchmarks, categorized into standard tasks (GSM8K and MATH500) and competition-level challenges (AMC23, AIME24, AIME25 and OlympiadBench) to assess varying degrees of reasoning depth.

We evaluate the proposed method and baselines on a comprehensive suite of mathematical reasoning benchmarks, categorized into standard tasks (GSM8K and MATH500) and competition-level challenges (AMC23, AIME24, AIME25, and OlympiadBench), along with a non-math scientific QA benchmark (GPQA-Diamond) to assess cross-domain generalization.

\begin{itemize}[leftmargin=*]
    \item \textbf{GSM8K} (Grade School Math 8K)~\cite{GSM8K}: A dataset consisting of 8.5K high-quality grade school math word problems. It requires models to perform multi-step elementary reasoning using basic arithmetic operations. We use the standard test set to evaluate the model's ability to handle linguistic diversity and sequential logic in relatively simple scenarios.
    \item \textbf{MATH500}~\cite{MATH}: A representative subset of the challenging MATH dataset, containing 500 problems sampled across seven categories (e.g., algebra, geometry, number theory) and five difficulty levels. MATH500 provides a computationally efficient yet rigorous estimate of a model's performance on broader pre-university competition mathematics.
    \item \textbf{AMC23}~\cite{AMC23}: Comprises problems from the 2023 American Mathematics Competitions (AMC 10 and AMC 12). These exams are the first tier in the United States high school mathematics competition series, requiring creative problem-solving skills beyond standard curriculum application.
    \item \textbf{AIME24 \& 25}~\cite{AIME}: The American Invitational Mathematics Examination is the second tier of the competition series, accessible only to top performers in the AMC. The problems in this set involve integers between 0 and 999 as answers and require significantly deeper reasoning and sustained logical chains compared to the AMC level.
    \item \textbf{OlympiadBench}~\cite{olympaid}: A comprehensive bilingual (English and Chinese) benchmark featuring Olympiad-level mathematics and physics problems. We focus on the mathematics subset, which sources problems from prestigious competitions such as the International Mathematical Olympiad (IMO) and the Chinese College Entrance Examination (GaoKao). This represents the upper bound of difficulty in our evaluation suite.
    \item \textbf{GPQA-Diamond}~\cite{GPQA}: A benchmark of graduate-level scientific question answering across biology, physics, and chemistry, written and validated by domain experts. The Diamond subset contains the most difficult ``Google-proof'' questions, requiring deep domain knowledge rather than retrieval. We use a 50-question subset to evaluate generalization beyond mathematical reasoning, with results reported in Appendix~\ref{apdx:gpqa_results}.
\end{itemize}

\clearpage
\section{Analysis Details} \label{apdx:Analysis}

\subsection{Lorenz Curve and Gini Coefficient for Token-Level Discrepancy}\label{apdx:lorenz_gini}

This section provides the formal definitions of the Lorenz curve and the Gini coefficient used in Section~\ref{subsec:sparsity}, as well as implementation details.

\paragraph{Token-level disagreement scores.}
For each response, we obtain a sequence of nonnegative per-token scores $\{s_t\}_{t=1}^N$ (e.g., cross-entropy discrepancy between base and reasoning next-token distributions).
We sort these scores in non-decreasing order:
$s_{(1)} \le s_{(2)} \le \cdots \le s_{(N)}$.

\paragraph{Lorenz curve.}~\cite{Lorenz}
The Lorenz curve is defined by points $(x_k, y_k)$ for $k=1,\ldots,N$, where
\begin{equation}
x_k = \frac{k}{N},
\qquad
y_k = \frac{\sum_{i=1}^k s_{(i)}}{\sum_{j=1}^N s_{(j)}}.
\end{equation}
Intuitively, $y_k$ measures what fraction of total discrepancy mass is contributed by the lowest-scoring $k$ tokens.
A curve close to the equality line $y=x$ indicates a nearly uniform contribution across tokens, while a strongly bowed curve indicates high concentration.

\paragraph{Gini coefficient.}~\cite{gini}
We summarize concentration by the Gini coefficient $G$, which equals twice the area between the Lorenz curve and the equality line:
\begin{equation}
G = 1 - 2\int_0^1 L(x)\,dx.
\end{equation}
For a discrete set of sorted scores, we compute
\begin{equation}
G = \frac{\sum_{i=1}^N (2i - N - 1)\, s_{(i)}}{N \sum_{j=1}^N s_{(j)}},
\end{equation}
which ranges from $0$ (perfectly uniform mass) to $1$ (all mass on a single token).

\paragraph{Aggregation across samples and datasets.}
We compute the Lorenz curve and Gini coefficient  {per sample} using that sample's token sequence, and then report the dataset-level mean (and optionally standard deviation / confidence intervals) across samples.
The overall value reported in Section~\ref{subsec:sparsity} is the average of dataset-level means across the six benchmarks.

\subsection{Token Enrichment Analysis: Definition and Token Categorization}
\label{apdx:enrichment_details}

Section~\ref{subsec:alignment} and Section~\ref{subsec:patterns} report two complementary token enrichment analyses~\cite{enrichment-analysis} using the same token categorization rules.
The first ( {peak-based enrichment}) is computed on base-model rollouts, measuring whether tokens at high-entropy or high cross-model discrepancy peaks are planning-heavy.
The second ( {takeover-based enrichment}) is computed on guided rollouts, measuring whether the tokens actually generated by the reasoning model during takeover are enriched for planning markers.
This appendix provides the shared enrichment definition and the token categorization procedure used in both analyses.

\paragraph{Enrichment ratio.}
Let $C$ be a token category (e.g., \textit{Planning}).
We compute an enrichment ratio
\begin{equation}
    E(C) \;=\; 
    \frac{P(t \in C \mid t \in \mathcal{T}_{\text{interv}}^{\text{sem}})}
         {P(t \in C \mid t \in \mathcal{T}_{\text{global}}^{\text{sem}})},
\end{equation}
where $\mathcal{T}_{\text{interv}}^{\text{sem}}$ denotes the set of  {intervened} tokens (i.e., tokens generated by the reasoning model during takeover) that are  {classifiable} into one of our semantic categories,
and $\mathcal{T}_{\text{global}}^{\text{sem}}$ denotes the set of  {classifiable} tokens in the standard base rollout.
Importantly, tokens that do not match any semantic rule (e.g., punctuation, stopwords, generic content words) are excluded from both the numerator/denominator counts and the normalizing totals, so the enrichment is computed  {within the planning/execution semantic subset}.

\paragraph{Token categorization (heuristic).}
We classify each decoded token $t$ into one of two categories:
(i) \textit{Planning/Decision} and (ii) \textit{Execution/Arithmetic}.
Classification uses a lightweight heuristic based on exact keyword matching (for planning) and regular-expression/symbol matching (for execution).
All comparisons are case-insensitive after stripping surrounding whitespace.

\paragraph{Planning / Decision Tokens.} This category includes words associated with reasoning structure, strategy selection, logical deduction, and self-reflection. We constructed a vocabulary $\mathcal{V}_{\text{plan}}$ containing several keywords across the following sub-categories:
\begin{itemize}[leftmargin=*]
    \item \textbf{Step Indicators:} \texttt{first}, \texttt{next}, \texttt{then}, \texttt{finally}, \texttt{step}.
    \item \textbf{Logical Connectives:} \texttt{therefore}, \texttt{so}, \texttt{since}, \texttt{because}, \texttt{thus}, \texttt{hence}, \texttt{implies}.
    \item \textbf{Strategy \& Action Verbs:} \texttt{let}, \texttt{assume}, \texttt{suppose}, \texttt{consider}, \texttt{calculate}, \texttt{solve}, \texttt{check}, \texttt{verify}, \texttt{notice}, \texttt{determine}, \texttt{choose}, \texttt{analyze}.
    \item \textbf{Contrast \& Correction:} \texttt{but}, \texttt{however}, \texttt{actually}, \texttt{wait}, \texttt{instead}, \texttt{alternatively}.
    \item \textbf{Thinking Markers:} \texttt{thinking}, \texttt{strategy}, \texttt{idea}, \texttt{plan}, \texttt{break down}, \texttt{approach}, \texttt{method}.
    \item \textbf{Conditionals \& Wh-Words:} \texttt{if}, \texttt{unless}, \texttt{whether}, \texttt{why}, \texttt{how}, \texttt{what}.
\end{itemize}

\paragraph{Execution / Arithmetic Tokens.} This category captures the mechanical execution of mathematical operations and symbolic manipulation. We identified these tokens using regular expressions:
\begin{itemize}[leftmargin=*]
    \item \textbf{Numbers:} Integers and floating-point numbers (e.g., \texttt{100}, \texttt{3.14}).
    \item \textbf{Operators:} Mathematical symbols including \texttt{+}, \texttt{-}, \texttt{*}, \texttt{/}, \texttt{=}, \texttt{<}, \texttt{>}, \texttt{\^}.
    \item \textbf{Structure:} LaTeX formatting symbols and brackets (e.g., \texttt{\$}, \texttt{\{}, \texttt{\}}, \texttt{(}, \texttt{)}).
    \item \textbf{Variables:} Single-letter variables common in math problems (e.g., \texttt{x}, \texttt{y}, \texttt{n}) when appearing in a mathematical context.
\end{itemize}

\paragraph{Target vs.\ reference token sets.}
Each enrichment analysis compares a target token set $\mathcal{T}_{\text{target}}$ to a  {reference} token set $\mathcal{T}_{\text{ref}}$, and computes category proportions within the classifiable semantic subset.
In the  {peak-based enrichment} (Section~\ref{subsec:alignment}), $\mathcal{T}_{\text{target}}$ is the top-$p\%$ tokens in a base rollout ranked by either base entropy $H_b(t)$ or cross-model discrepancy $s_t$, and $\mathcal{T}_{\text{ref}}$ is the full base-rollout token stream.
In the  {takeover-based enrichment} (Section~\ref{subsec:patterns}), $\mathcal{T}_{\text{target}}$ consists of takeover tokens extracted from guided rollouts using the per-step provenance field \texttt{chosen\_from} (sources beginning with \texttt{teacher}), and $\mathcal{T}_{\text{ref}}$ is the corresponding base-rollout token stream.

\paragraph{Interpretation and limitations.}
This enrichment analysis is intended as a coarse, interpretable proxy for identifying planning markers.
While it does not capture all planning behavior (e.g., implicit planning without explicit markers), the strong enrichment of planning tokens in the intervention set supports the claim that our takeover mechanism disproportionately targets planning-related linguistic cues rather than execution-heavy arithmetic tokens.

\subsection{Per-dataset planning enrichment at entropy/disagreement peaks}
\label{apdx:base_planning_enrichment_per_dataset}

\begin{table}[ht]
    \centering
    \caption{\textbf{Planning enrichment at peaks across datasets (top-1\%).}
    For each dataset, we report the planning share in the base-rollout global token stream and among the top-1\% tokens selected by either cross-model divergence $s_t$ (CE) or base entropy $H_b(t)$, together with the enrichment ratio.}
    \label{tab:base_planning_enrichment_per_dataset}
    \setlength{\tabcolsep}{2.5pt}
    \begin{tabular}{lccc|ccc}
        \toprule
        & \multicolumn{3}{c}{Disagreement $s_t$ (CE)} & \multicolumn{3}{c}{Entropy $H_b(t)$} \\
        \cmidrule(lr){2-4}\cmidrule(lr){5-7}
        Dataset & Global (\%) & Top-1\% (\%) & Enrich. & Global (\%) & Top-1\% (\%) & Enrich. \\
        \midrule
        Math500 & 2.22 & 15.98 & 7.21$\times$ & 2.22 & 15.90 & 7.17$\times$ \\
        GSM8K & 1.42 & 5.19 & 3.65$\times$ & 1.42 & 7.39 & 5.19$\times$ \\
        AMC23 & 1.63 & 16.37 & 10.05$\times$ & 1.63 & 17.94 & 11.01$\times$ \\
        OlympiadBench & 1.92 & 14.19 & 7.41$\times$ & 1.92 & 14.53 & 7.59$\times$ \\
        AIME24 & 1.96 & 15.41 & 7.86$\times$ & 1.96 & 13.28 & 6.77$\times$ \\
        AIME25 & 1.72 & 29.18 & 16.98$\times$ & 1.72 & 34.95 & 20.34$\times$ \\
        \bottomrule
    \end{tabular}
\end{table}

Table~\ref{tab:base_planning_enrichment} aggregates planning enrichment across all datasets.
Here we provide the per-dataset breakdown to assess consistency and heterogeneity across benchmarks.
Table~\ref{tab:base_planning_enrichment_per_dataset} shows that planning enrichment at top-1\% peaks holds across all datasets for both ranking signals, with particularly large effects on harder benchmarks (e.g., AMC23 and AIME25).

\clearpage
\section{Additional Results} \label{apdx:add_results}

\subsection{Reverse-KL as an Alternative Disagreement Signal}
\label{apdx:rkl_results}

Our method is not tied to Cross Entropy as the disagreement metric.
We replicate the gated takeover using  {reverse KL}, i.e., $s_t=\mathrm{KL}(p_b\|p_r)$, while keeping the same calibration protocol (global spike ratio $r$ and sliding-window smoothing with $W=64$) and the same evaluation setup.
Table~\ref{tab:main_performance_rkl} reports the results and shows that reverse-KL yields strong recovery under comparable realized budgets, confirming that divergence-guided takeover is not specific to a single discrepancy function.

\begin{table}[h]
    \centering
    \small
    \setlength{\tabcolsep}{3pt}
    \caption{Experimental results on mathematical reasoning benchmarks (Accuracy / Pass@8). The Recovery column quantifies how much the intervention recovers reasoning capabilities, calculated as $(P_{Guided} - P_{Base}) / (P_{Thinking} - P_{Base})$, where $P$ denotes Pass@8 (consistent with Table~\ref{tab:main_performance}).}
    \label{tab:main_performance_rkl}
    \begin{tabular}{l ccccccc c}
        \toprule
        Model & Math500 & GSM8K & AMC23 & Olympiad & AIME24 & AIME25 & Average & Recovery\\
        \midrule
        Qwen3-0.6B-Base        & 27.6 / 67.0 & 29.9 / 82.0 & 13.4 / 42.5 & 6.8 / 21.0 & 0.4 / 3.3   & 0.0 / 0.0   & 13.0 / 36.0 & - \\
        \rowcolor{lightred}
        ~ + Guided ($\bar{\rho} \approx 0.08$) & 82.9 / 99.0 & 87.4 / 98.0 & 64.4 / 92.5 & 42.4 / 58.0 & 27.9 / 53.3 & 27.5 / 46.7 & 55.4 / 74.6 & $137\%$ \\
        \rowcolor{lightred}
        ~ + Guided ($\bar{\rho} \approx 0.17$) & 90.6 / 100.0 & 93.8 / 99.0 & 84.1 / 97.5 & 47.9 / 62.0 & 40.8 / 60.0 & 35.0 / 60.0 & 65.4 / 79.8 & $156\%$ \\
        \rowcolor{lightgray}
        Qwen3-0.6B (Thinking)  & 72.1 / 90.0 & 81.6 / 94.0 & 46.9 / 80.0 & 34.3 / 54.0 & 9.2 / 30.0  & 16.1 / 36.7 & 43.4 / 64.1 & - \\
        \midrule
        Qwen3-8B (Thinking)   & 97.4 / 100.0 & 97.8 / 98.0 & 90.9 / 100.0 & 60.0 / 69.0 & 70.0 / 83.3 & 52.5 / 73.3 & 78.1 / 87.3 & - \\
        \bottomrule
    \end{tabular}
\end{table}

\begin{table}[h]
    \centering
    \small
    \setlength{\tabcolsep}{5pt}
    \caption{\textbf{Recovery under different disagreement metrics (Qwen3-0.6B).} 
    We report Recovery computed on Accuracy and Pass@8 respectively.}
    \label{tab:recovery_metric_compare}
    \begin{tabular}{lccc}
        \toprule
        Metric & Budget & Recovery (Acc) & Recovery (Pass@8) \\
        \midrule
        CE   & $\bar{\rho}\!\approx\!0.04$ & 53\%  & 91\%  \\
             & $\bar{\rho}\!\approx\!0.13$ & 130\% & 157\% \\
        \midrule
        rKL  & $\bar{\rho}\!\approx\!0.08$ & 140\% & 137\% \\
             & $\bar{\rho}\!\approx\!0.17$ & 172\% & 156\% \\
        \bottomrule
    \end{tabular}
\end{table}

\begin{figure}[h]
    \centering
    \begin{subfigure}[t]{0.49\linewidth}
        \centering
        
        \includegraphics[width=\linewidth]{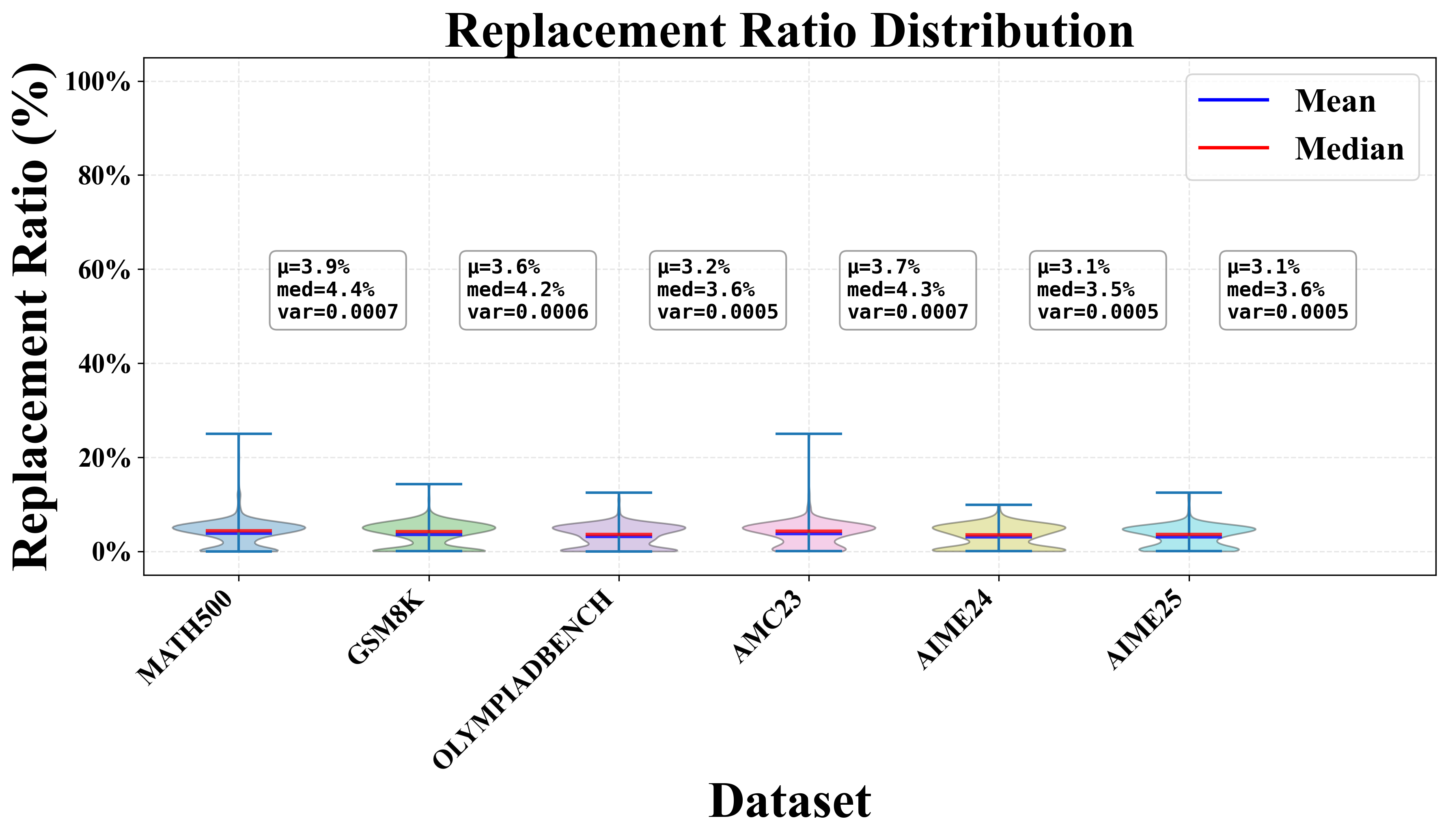} 
        \caption{CE gate ($r=0.03$)}
        \label{fig:rho_violin_ce_003}
    \end{subfigure}
    \hfill
    \begin{subfigure}[t]{0.49\linewidth}
        \centering
        
        \includegraphics[width=\linewidth]{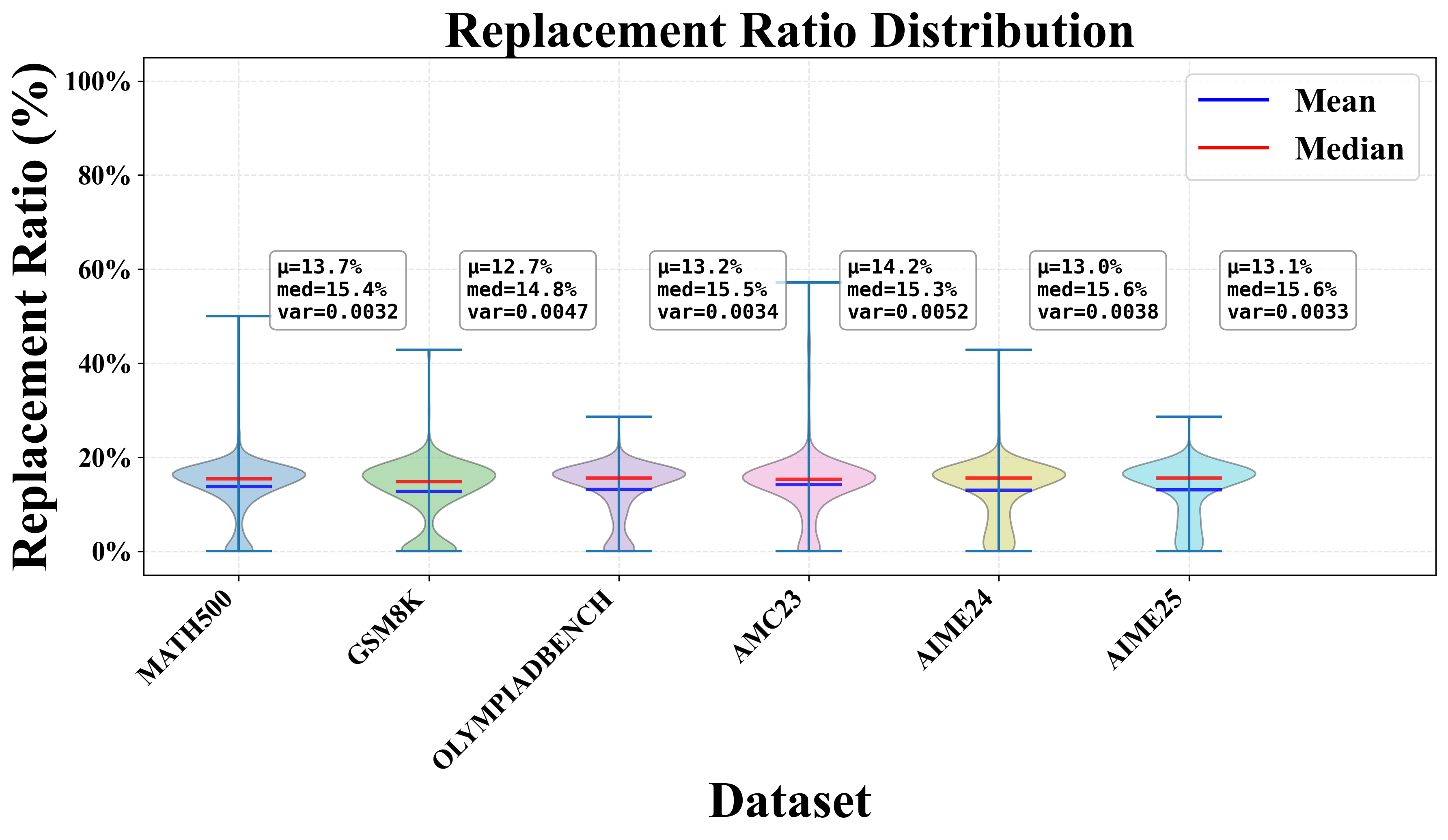}
        \caption{CE gate ($r=0.05$)}
        \label{fig:rho_violin_ce_005}
    \end{subfigure}

    \vspace{4pt}

    \begin{subfigure}[t]{0.49\linewidth}
        \centering
        
        \includegraphics[width=\linewidth]{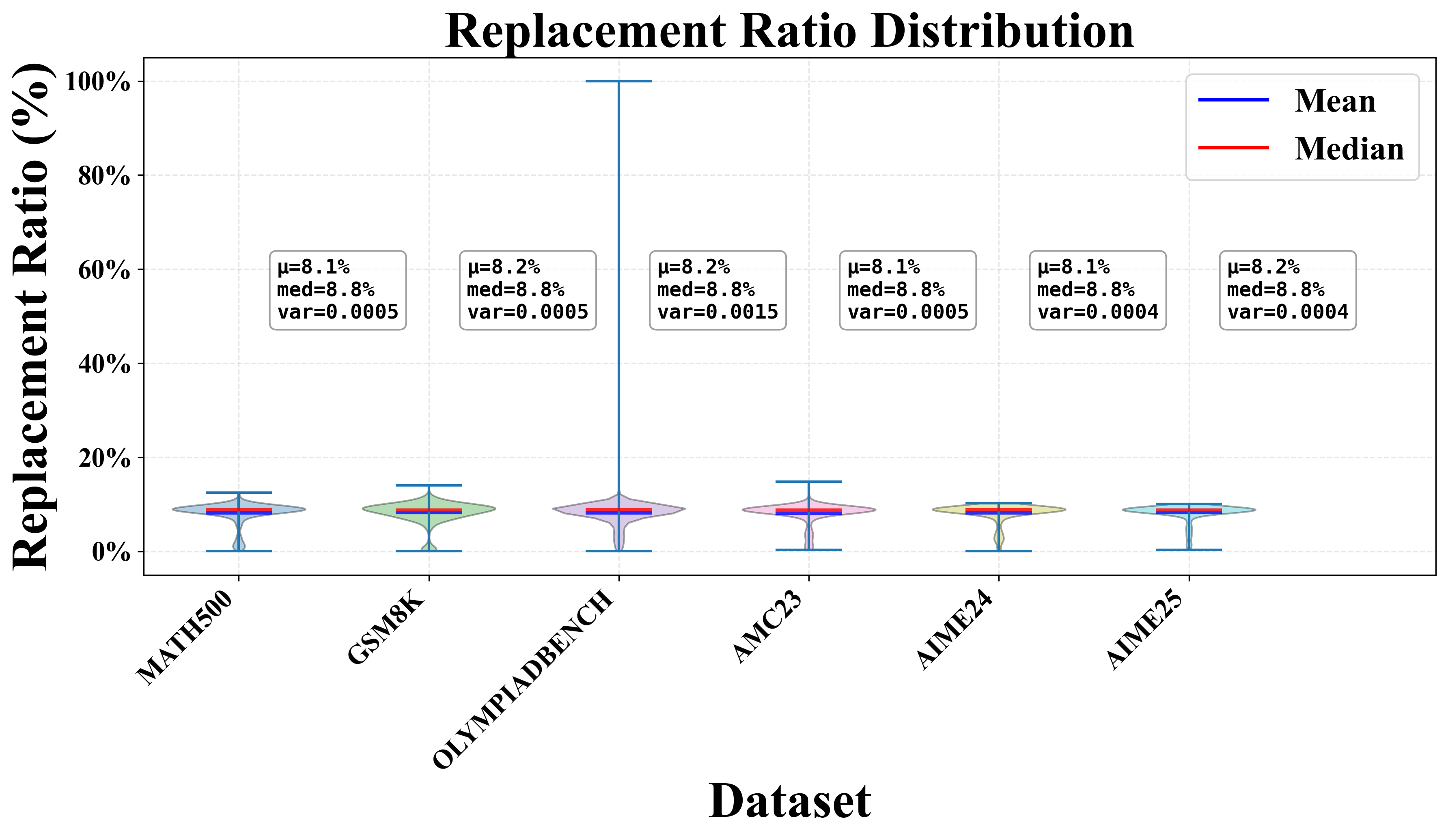}
        \caption{rKL gate ($r=0.03$)}
        \label{fig:rho_violin_rkl_003}
    \end{subfigure}
    \hfill
    \begin{subfigure}[t]{0.49\linewidth}
        \centering

        \includegraphics[width=\linewidth]{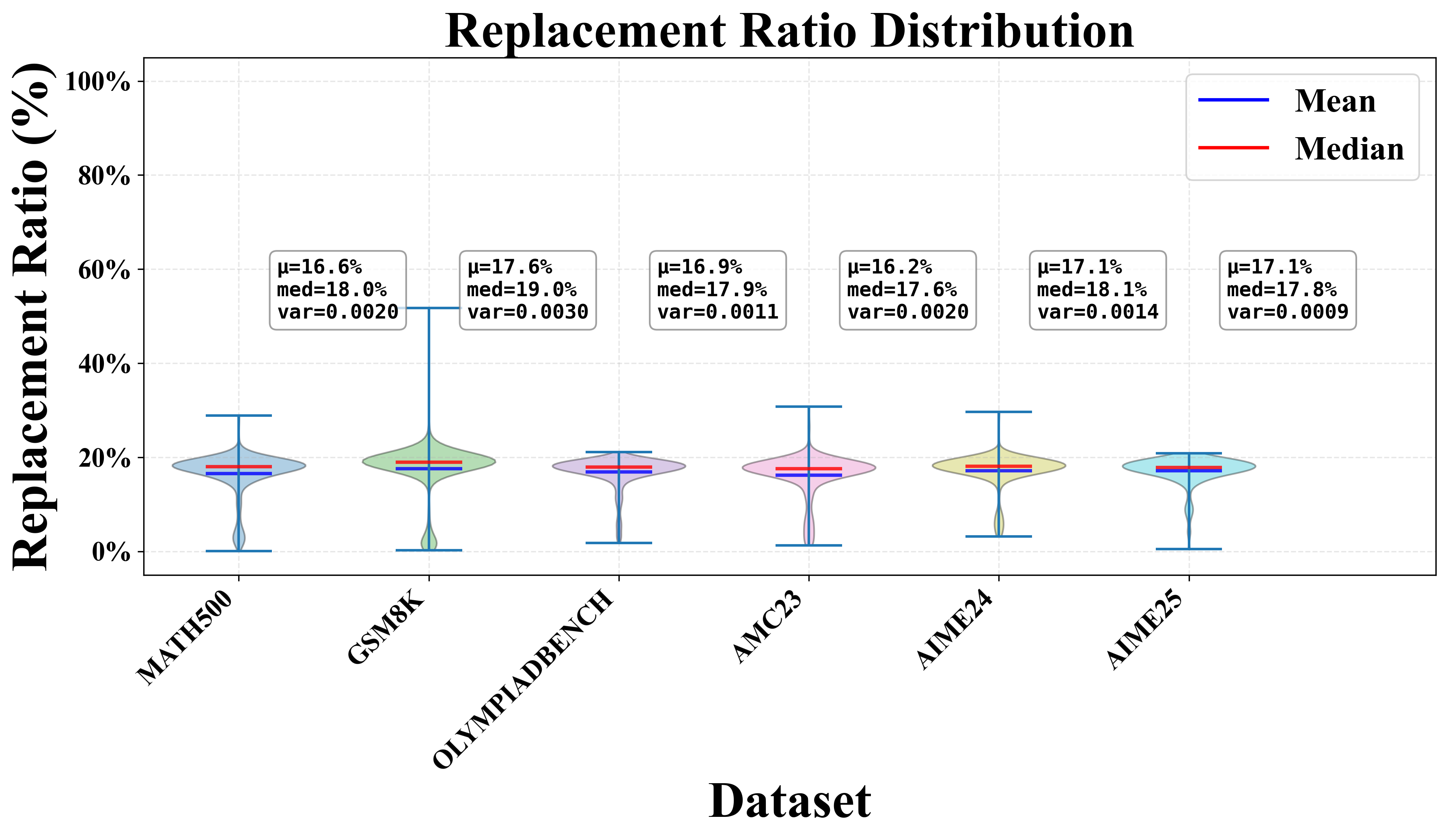}
        \caption{rKL gate ($r=0.05$)}
        \label{fig:rho_violin_rkl_005}
    \end{subfigure}

    \caption{\textbf{Stability of realized replacement budget.} Violin plots of per-problem replacement rate $\rho_i$ across benchmarks for Qwen3-0.6B under (top) Cross-Entropy gating and (bottom) reverse-KL gating, each shown at two calibrated spike ratios ($r=0.03$ and $r=0.05$). Concentrated distributions indicate that the realized intervention budget is stable across problems rather than dominated by a small subset of outliers.}
    \label{fig:rho_violin}
\end{figure}

To better understand metric-dependent behavior, Table~\ref{tab:recovery_metric_compare} summarizes Recovery computed on Accuracy and on Pass@8, respectively.
Empirically, reverse-KL tends to yield higher Accuracy-based recovery, whereas Cross Entropy yields stronger Pass@8-based recovery.
A plausible explanation is that $\mathrm{KL}(p_b\|p_r)$ emphasizes cases where the base model places overly concentrated mass on an incorrect token (confidence miscalibration), while Cross Entropy more directly reflects distributional mismatch and can better improve the probability that  {at least one} sampled trajectory reaches the correct solution.

\subsection{Stability of the Realized Replacement Budget}
\label{apdx:budget_stability}

We examine the per-problem realized replacement rate $\rho_i=\frac{1}{T_i}\sum_{t=1}^{T_i} g_t$.
Figure~\ref{fig:rho_violin} visualizes the distribution of $\rho_i$ across benchmarks for Qwen3-0.6B under Cross-Entropy and reverse-KL gating, each at two calibrated spike ratios ($r=0.03$ and $r=0.05$).

Two observations are consistent across datasets.
First, the distributions are concentrated with small variance, indicating that the realized intervention budget is stable across problems rather than dominated by a small subset of outliers.
Second, for the same $r$, Cross Entropy consistently yields a  {lower} realized replacement rate than reverse-KL, suggesting that CE produces a more selective gate under our calibration procedure.
This motivates reporting budgets using realized $\rho$ (and using $\bar{\rho}$ in tables) when comparing metrics and settings.

\subsection{Entropy-Triggered Practical Variant}\label{apdx:entropy_variant}

The disagreement gate in \cref{sec:intervention} queries both $p_b$ and $p_r$ at every step, which is fine for diagnosis but wasteful at deployment. Since disagreement spikes overlap heavily with base-entropy spikes (\cref{subsec:alignment}, Figure~\ref{fig:iou}), a cheaper alternative is to trigger on the base entropy $H_b(t)=H(p_b(\cdot\!\mid\!x_t))$ alone, applying the same calibration and gating recipe with $s_t$ replaced by $H_b(t)$. As shown in Table~\ref{tab:entropy_guided}, this cheaper trigger already recovers most of the strong-model Pass@8 advantage: at $\bar{\rho}\!\approx\!0.08$, the entropy-guided variant reaches $65.8/87.2$ (Acc/Pass@8), essentially matching Qwen3-8B's Pass@8 of $87.3$ while querying the stronger model at only $\sim$8\% of steps.

\begin{table}[h]
    \centering
    \small
    \setlength{\tabcolsep}{4pt}
    \caption{\textbf{Entropy-triggered practical variant (Qwen3-0.6B).}
    Replacing the cross-model disagreement trigger with base-model entropy $H_b(t)$ already recovers most of the strong-model Pass@8 advantage while only requiring $p_b$ at every step.}
    \label{tab:entropy_guided}
    \begin{tabular}{lcc}
        \toprule
        Method & Avg ACC & Avg Pass@8 \\
        \midrule
        Qwen3-0.6B-Base                                       & 12.9 & 36.0 \\
        \rowcolor{lightred}
        ~ + Entropy-Guided ($\bar{\rho}\!\approx\!0.08$)      & 65.8 & 87.2 \\
        \rowcolor{lightred}
        ~ + Entropy-Guided ($\bar{\rho}\!\approx\!0.20$)      & 71.3 & 86.4 \\
        \rowcolor{lightgray}
        Qwen3-0.6B (Thinking)                                 & 43.4 & 64.1 \\
        \rowcolor{lightgray}
        Qwen3-8B (Thinking)                                   & 78.1 & 87.3 \\
        \bottomrule
    \end{tabular}
\end{table}

\subsection{Matched-Budget Baselines: Random and Early-Only}\label{apdx:baselines}

We test whether the gains from disagreement-guided intervention are driven specifically by the choice of \scs{which} tokens are replaced, by comparing against two budget-matched baselines:
(i) a \textbf{random} baseline that uniformly samples replacement positions to hit a target budget $\rho$, and
(ii) an \textbf{early-only} baseline that, given the reasoning model's output of length $L_r$, replaces the first $\rho\!\cdot\! L_r$ tokens of the rollout with reasoning-model tokens and then lets the base model continue.
The early-only baseline is motivated by our Finding 2 (\cref{subsec:structure}) that disagreement spikes concentrate at early positions: it isolates whether \scs{position alone} is sufficient.

Table~\ref{tab:matched_budget_baselines} summarizes the comparison on Qwen3-0.6B (averaged over the six math benchmarks).
Two findings stand out:

\textbf{Random replacement is consistently dominated by guided.}
At an aggressive $25\%$ random replacement, average performance only reaches $26.4/55.2$ (Acc/Pass@8), whereas the disagreement-guided gate at only $\bar{\rho}\!\approx\!0.04$ already reaches $29.1/61.4$.
This confirms that gains do not come merely from inserting stronger-model tokens at arbitrary positions.

\textbf{Position alone is not sufficient.}
The early-only baseline does improve over the base, consistent with our finding that critical tokens concentrate early.
However, even at $\bar{\rho}\!\approx\!0.25$ early-only reaches only $25.7/58.4$, which is still worse than guided at $\bar{\rho}\!\approx\!0.04$ ($29.1/61.4$).
This shows that within the early region, only the disagreement / uncertainty spikes are truly trajectory-setting; blindly replacing all early tokens wastes budget on non-critical positions.

Together, these baselines support the interpretation that the gain comes from selecting positions that are simultaneously \scs{early} and \scs{informative}, i.e., positions where the base and reasoning models disagree most under base-model uncertainty.

\begin{table}[h]
    \centering
    \small
    \setlength{\tabcolsep}{5pt}
    \caption{\textbf{Matched-budget baselines on Qwen3-0.6B.}
    Average Accuracy / Pass@8 over the six math reasoning benchmarks.
    Random samples replacement positions uniformly under a target budget $\rho$.
    Early-only replaces the first $\rho\!\cdot\!L_r$ tokens (where $L_r$ is the reasoning-model rollout length) and then lets the base model continue.
    Disagreement-guided intervention at $\bar{\rho}\!\approx\!0.04$ already outperforms both baselines at $25\%$ budget.}
    \label{tab:matched_budget_baselines}
    \begin{tabular}{lccc}
        \toprule
        Method & Budget $\rho$ & Avg ACC & Avg Pass@8 \\
        \midrule
        Qwen3-0.6B-Base                  & 0.00                       & 13.0  & 36.0  \\
        \midrule
        ~ + Random                       & 0.10                       & 17.7 & 42.0 \\
        ~ + Random                       & 0.15                       & 20.1 & 46.0 \\
        ~ + Random                       & 0.20                       & 21.4 & 50.0 \\
        ~ + Random                       & 0.25                       & 26.4 & 55.2 \\
        \midrule
        ~ + Early-only                   & 0.10                       & 20.9 & 44.1 \\
        ~ + Early-only                   & 0.15                       & 20.9 & 49.9 \\
        ~ + Early-only                   & 0.20                       & 23.3 & 53.4 \\
        ~ + Early-only                   & 0.25                       & 25.7 & 58.4 \\
        \midrule
        \rowcolor{lightred}
        ~ + Guided (Ours)                & $\bar{\rho}\!\approx\!0.04$ & 29.1 & 61.4 \\
        \rowcolor{lightred}
        ~ + Guided (Ours)                & $\bar{\rho}\!\approx\!0.13$ & 52.4 & 80.0 \\
        \midrule
        \rowcolor{lightgray}
        Qwen3-0.6B (Thinking)            & --                         & 43.4 & 64.1 \\
        \rowcolor{lightgray}
        Qwen3-8B (Thinking)              & --                         & 78.1 & 87.3 \\
        \bottomrule
    \end{tabular}
\end{table}

\subsection{Cross-Family Generalization on LLaMA}\label{apdx:llama_results}

To test whether the sparse-control finding transfers beyond the Qwen3 family, we evaluate the disagreement-guided intervention on a LLaMA-based base/guide pair: the base model is LLaMA-3.1-8B~\cite{Llama3} and the reasoning guide is DeepSeek-R1-Distill-Llama-8B~\cite{deepseek-r1}.
Calibration follows the same procedure as in the main experiments (held-out GSM8K-style prompts, sliding window $W=64$).

Table~\ref{tab:llama_main} reports the average performance over the six math benchmarks.
The same monotonic recovery pattern holds: with $\bar{\rho}\!\approx\!0.20$, guided intervention recovers $\sim$91\% of the gap (in Pass@8) between the base model and the reasoning guide, mirroring the Qwen3 results in Table~\ref{tab:main_performance}.
This indicates that the disagreement-spike phenomenon and the resulting recovery curve are not specific to a single model family or training pipeline.

\begin{table}[h]
    \centering
    \small
    \setlength{\tabcolsep}{3pt}
    \caption{\textbf{Cross-family generalization on the LLaMA pair (Accuracy / Pass@8).}
    Base model: LLaMA-3.1-8B; reasoning guide: DeepSeek-R1-Distill-Llama-8B.
    The Recovery column quantifies how much the intervention recovers reasoning capabilities, calculated as $(P_{Guided} - P_{Base}) / (P_{Guide} - P_{Base})$, where $P$ denotes Pass@8 (consistent with Table~\ref{tab:main_performance}).}
    \label{tab:llama_main}
    \begin{tabular}{l ccccccc c}
        \toprule
        Model & Math500 & GSM8K & AMC23 & Olympiad & AIME24 & AIME25 & Average & Recovery\\
        \midrule
        LLaMA-3.1-8B (Base)
            & 6.0 / 30.0
            & 16.9 / 66.0
            & 2.8 / 17.5
            & 1.1 / 7.0
            & 0.8 / 6.7
            & 0.0 / 0.0
            & 4.6 / 21.2
            & - \\
        \rowcolor{lightred}
        ~ + Guided ($\bar{\rho}\!\approx\!0.03$)
            & 19.4 / 64.0
            & 41.8 / 87.0
            & 6.2 / 30.0
            & 3.2 / 12.0
            & 1.2 / 6.7
            & 0.4 / 3.3
            & 12.0 / 33.8
            & $21\%$ \\
        \rowcolor{lightred}
        ~ + Guided ($\bar{\rho}\!\approx\!0.12$)
            & 55.5 / 90.0
            & 70.9 / 97.0
            & 49.4 / 90.0
            & 26.2 / 54.0
            & 14.2 / 43.3
            & 11.7 / 33.3
            & 38.0 / 67.9
            & $78\%$ \\
        \rowcolor{lightred}
        ~ + Guided ($\bar{\rho}\!\approx\!0.20$)
            & 74.6 / 97.0
            & 86.2 / 97.0
            & 72.5 / 92.5
            & 41.2 / 60.0
            & 35.4 / 66.7
            & 22.5 / 43.3
            & 55.4 / 76.1
            & $91\%$ \\
        \rowcolor{lightgray}
        DeepSeek-R1-Distill-Llama-8B (Guide)
            & 86.9 / 98.0
            & 90.6 / 97.0
            & 86.2 / 95.0
            & 51.9 / 65.0
            & 42.9 / 80.0
            & 30.0 / 53.3
            & 64.8 / 81.4
            & - \\
        \bottomrule
    \end{tabular}
\end{table}

\subsection{Cross-Domain Generalization to Non-math Scientific QA}\label{apdx:gpqa_results}

To assess whether the sparse-intervention effect transfers beyond mathematical reasoning, we evaluate on GPQA-Diamond~\cite{GPQA}, a benchmark of graduate-level scientific question answering across biology, physics, and chemistry. We use a 50-question subset and keep the same Qwen3-0.6B-Base / Qwen3-8B (Thinking) pair as in our main experiments.

Table~\ref{tab:gpqa_results} shows two patterns that differ from the math setting. First, same-size reasoning post-training provides little advantage here: Qwen3-0.6B (Thinking) matches the base in Accuracy ($23.0$ vs.\ $22.5$) and \scs{underperforms} it in Pass@8 ($68.0$ vs.\ $78.0$); even the 8B reasoning model does not improve Pass@8 over the 0.6B base (both $78.0$). Second, sparse intervention nonetheless yields clear gains: at $\bar{\rho}\!\approx\!0.05$, guided decoding reaches $37.1\,/\,88.0$, improving the base by $+14.6$ Accuracy and $+10.0$ Pass@8, and exceeding the 8B reasoning model's Pass@8 ($88.0$ vs.\ $78.0$). The remaining Accuracy gap to 8B ($37.1$ vs.\ $63.2$) is larger than on math benchmarks, suggesting that fully closing the non-math reasoning gap may require denser intervention. Together with the cross-family LLaMA results in Appendix~\ref{apdx:llama_results}, this indicates the sparse-control phenomenon is not specific to a single model family or to mathematical reasoning.

\begin{table}[h]
    \centering
    \small
    \setlength{\tabcolsep}{8pt}
    \caption{\textbf{Cross-domain generalization on GPQA-Diamond (50-question subset; Accuracy / Pass@8).}
    Base: Qwen3-0.6B-Base; reasoning guide: Qwen3-8B (Thinking). Same-size reasoning post-training provides little Pass@8 advantage in this domain, while sparse guided intervention improves both Accuracy and Pass@8 over the base and exceeds the 8B reasoning model's Pass@8.}
    \label{tab:gpqa_results}
    \begin{tabular}{lcc}
        \toprule
        Model & Accuracy & Pass@8 \\
        \midrule
        Qwen3-0.6B-Base                       & 22.5 & 78.0 \\
        Qwen3-0.6B (Thinking)                 & 23.0 & 68.0 \\
        \midrule
        \rowcolor{lightred}
        ~ + Guided ($\bar{\rho}\!\approx\!0.01$) & 18.8 & 82.0 \\
        \rowcolor{lightred}
        ~ + Guided ($\bar{\rho}\!\approx\!0.03$) & 26.1 & 78.0 \\
        \rowcolor{lightred}
        ~ + Guided ($\bar{\rho}\!\approx\!0.05$) & \textbf{37.1} & \textbf{88.0} \\
        \midrule
        \rowcolor{lightgray}
        Qwen3-8B (Thinking)                   & 63.2 & 78.0 \\
        \bottomrule
    \end{tabular}
\end{table}

\subsection{Calibration Threshold Stability across Datasets}\label{apdx:calibration_sensitivity}

A natural concern with calibrated gating is whether the calibrated thresholds $\tau$ and the tail-to-mean factor $\lambda$ (\cref{eq:lambda_def}) are dataset-specific or transferable.
To assess this, we re-calibrate the gate on each of the six benchmarks separately and compare the resulting $(\tau,\lambda)$ at three target spike ratios $r\in\{0.01,0.03,0.05\}$.
Table~\ref{tab:calibration_sensitivity} shows that:
(i) for a fixed $r$, the calibrated $\tau$ values stay within a narrow range across calibration sets (e.g., for $r=0.01$, $\tau\in[3.18, 4.12]$), and
(ii) both $\tau$ and $\lambda$ follow the same monotonic decreasing trend in $r$ across all datasets.
This indicates that the disagreement tail used by our gate is reasonably stable, rather than strongly dataset-specific, which is consistent with the calibration recipe used throughout the main paper (a single calibration set is used for all evaluations).

\begin{table}[h]
    \centering
    \small
    \setlength{\tabcolsep}{4pt}
    \caption{\textbf{Calibration threshold stability across datasets.}
    For each calibration set we report $\tau\,/\,\lambda$ at three target spike ratios $r\in\{0.01,0.03,0.05\}$.
    Values stay within a narrow range across datasets at fixed $r$, and follow the same monotonic trend with $r$, suggesting that the calibrated tail of the disagreement distribution is stable rather than dataset-specific.}
    \label{tab:calibration_sensitivity}
    \begin{tabular}{lccc}
        \toprule
        Calibration set & $\tau\,/\,\lambda$ ($r=0.01$) & $\tau\,/\,\lambda$ ($r=0.03$) & $\tau\,/\,\lambda$ ($r=0.05$) \\
        \midrule
        MATH500          & 3.649 / 23.378 & 1.095 / 7.018 & 0.372 / 2.381 \\
        GSM8K            & 3.667 / 24.443 & 0.467 / 3.114 & 0.239 / 1.591 \\
        OlympiadBench    & 3.509 / 24.046 & 0.997 / 6.834 & 0.293 / 2.011 \\
        AMC23            & 3.630 / 19.675 & 1.164 / 6.308 & 0.539 / 2.919 \\
        AIME24           & 3.181 / 27.816 & 0.772 / 6.751 & 0.187 / 1.635 \\
        AIME25           & 4.121 / 23.361 & 1.306 / 7.406 & 0.417 / 2.363 \\
        \bottomrule
    \end{tabular}
\end{table}

\subsection{Intervention Harm Analysis (Pass@8 Flip Statistics)}\label{apdx:harm_analysis}

A natural concern is whether the intervention, while improving aggregate performance, can also degrade specific samples that the base model would have answered correctly without help.
To quantify this, we examine sample-level Pass@8 label flips between the base model and the guided model on Qwen3-0.6B (with Qwen3-8B as the guide), at $\bar{\rho}\!\approx\!0.13$.
For each problem we compare whether \scs{any} of the 8 sampled trajectories under the base setting and under the guided setting are correct, and we tally three outcomes:
\textbf{Error$\to$Correct} (the guided setting now has at least one correct sample, while the base did not),
\textbf{Correct$\to$Error} (the base had at least one correct sample, but the guided does not),
and unchanged (both correct or both incorrect; not shown in the table).

\begin{table}[h]
    \centering
    \small
    \setlength{\tabcolsep}{6pt}
    \caption{\textbf{Sample-level Pass@8 flips on Qwen3-0.6B (guided vs.\ base, $\bar{\rho}\!\approx\!0.13$).}
    Across $400$ examples, the intervention overwhelmingly converts errors to successes (152 flips), while only 3 cases flip in the opposite direction.}
    \label{tab:harm_analysis}
    \begin{tabular}{lccc}
        \toprule
        Dataset & \#Examples & Error$\to$Correct & Correct$\to$Error \\
        \midrule
        AIME24         & 30  & 20  & 0 \\
        AIME25         & 30  & 17  & 0 \\
        AMC23          & 40  & 20  & 1 \\
        GSM8K          & 100 & 18  & 1 \\
        MATH500        & 100 & 32  & 0 \\
        OlympiadBench  & 100 & 45  & 1 \\
        \midrule
        \textbf{Total} & 400 & \textbf{152} & \textbf{3} \\
        \bottomrule
    \end{tabular}
\end{table}

Table~\ref{tab:harm_analysis} shows that the dominant effect of intervention is error correction: out of 400 examples, $152$ examples transition from incorrect to correct under guided decoding, while only $3$ transition in the opposite direction (a $\sim$50:1 ratio).
The three Correct$\to$Error cases are spread across AMC23, GSM8K, and OlympiadBench, and we observe no systematic pattern: each is a case where the base model had a borderline-correct trajectory that the intervention diverted toward a different (incorrect) chain.
This tail behavior is consistent with the fact that intervention edits the trajectory at a small set of high-uncertainty decision points: in rare cases, redirecting these decisions can move a borderline-correct rollout off-track.

\clearpage
\section{Case Study} \label{apdx:case_study}

\begin{figure*}[t]
    \centering
    \includegraphics[width=0.98\textwidth]{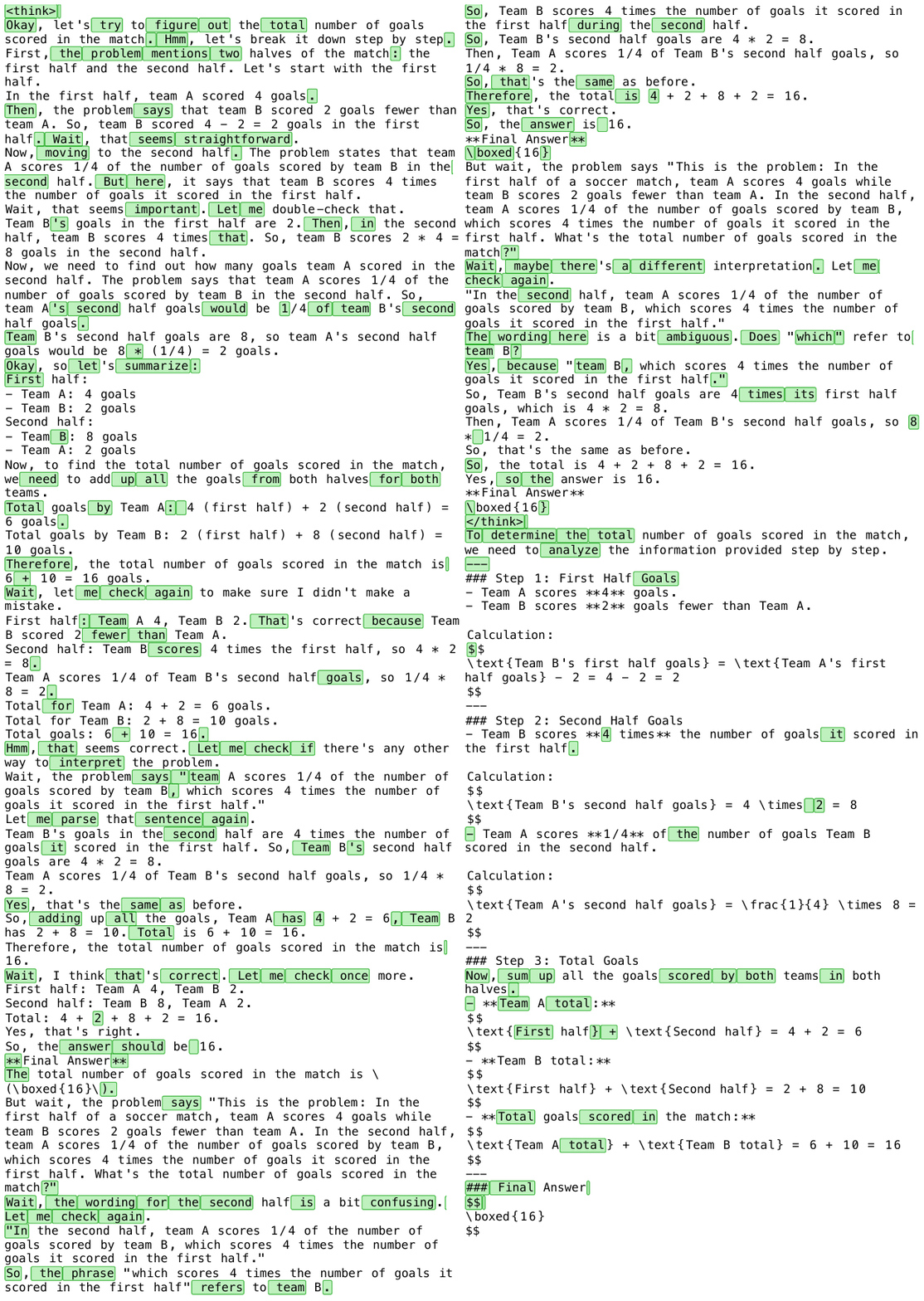}
    \caption{\textbf{Case study rollout.} Highlighted tokens denote reasoning model replacements triggered by disagreement spikes.}
    \label{fig:apdx-case-rollout}
\end{figure*}

We visualize the divergence-guided takeover on this example.
Figure~\ref{fig:apdx-case-problem} shows the prompt, and Figure~\ref{fig:apdx-case-rollout} provides the full guided rollout.

\paragraph{What gets replaced.}
A salient pattern is that most replaced tokens occur in planning- and control-related regions of the response: the brief outline of the solution, the choice of intermediate variables, and the decision of which quantities to compute next.
In contrast, once the correct plan is established, the subsequent arithmetic manipulations and execution steps are largely produced by the base model itself, with few (or no) additional takeovers.

\paragraph{Interpretation.}
This example supports the view that the base--reasoning gap is not primarily due to missing low-level execution capability.
Rather, failures concentrate at a small number of trajectory-defining decisions---what to do next and how to structure the computation---where the base model tends to commit to an unproductive path.
By selectively overriding these planning tokens, the takeover effectively supplies a lightweight ``controller'' that keeps the rollout on-track, while leaving the base model to carry out the bulk of the calculation.

\paragraph{Connection to our main findings.}
Consistent with our sparsity analysis, the interventions are localized and concentrated, yet have outsized downstream effects: correcting a few high-disagreement planning choices is sufficient to steer the remainder of the trajectory toward a correct solution.

\begin{figure}[h]
    \centering
    \begin{minipage}{0.98\linewidth}
        \small
        \begin{problembox}
        \textbf{Problem:} In the first half of a soccer match, team A scores 4 goals while team B scores 2 goals fewer than team A. In the second half, team A scores 1/4 of the number of goals scored by team B, which scores 4 times the number of goals it scored in the first half. What's the total number of goals scored in the match?
        \end{problembox}
    \end{minipage}
    \caption{\textbf{Case study prompt.} The following figure visualize the token-level takeover trajectory for this example.}
    \label{fig:apdx-case-problem}
\end{figure}

\end{document}